\documentclass{article}

\usepackage{graphicx}
\usepackage{wrapfig}
\usepackage{multirow}
\usepackage{enumitem}
\usepackage{booktabs}
\usepackage{xcolor}
\usepackage{tcolorbox}
\usepackage{subcaption}

 \usepackage[preprint]{neurips_2026}

\usepackage[utf8]{inputenc} 
\usepackage[T1]{fontenc}    
\usepackage{hyperref}       
\usepackage{url}            
\usepackage{booktabs}       
\usepackage{amsfonts}       
\usepackage{nicefrac}       
\usepackage{microtype}      
\usepackage{xcolor}         

\title{Actionable Hallucination Detection: Translating Latent Uncertainty into Agentic Critique}

\usepackage{authblk}
\author[1]{\textbf{Sanidhya Vijayvargiya}\textsuperscript{*}}
\author[1]{\textbf{Rahul Lokesh}}
\affil[1]{Samsung Research America}
\affil[ ]{\texttt{sanidhya.v@samsung.com, rahul.lokesh@samsung.com}}

\makeatletter
\renewcommand\AB@affilsepx{, \protect\Affilfont}
\makeatother

\begin{document}

\maketitle

\begin{abstract}
Large Language Models (LLMs) deployed as AI agents frequently exhibit user specification-grounding failures, executing hallucinated, undesired actions to force a resolution rather than expressing uncertainty. Existing detection methods fail to provide actionable, real-time correction as they either do not localize the hallucinations, or incur prohibitive inference latency. We introduce the Latent Critic, a lightweight low-rank adapter (LoRA) that operates concurrently with a frozen base LLM's generation to actively restructure the transformer's residual stream---amplifying latent grounding signals and translating them into localized, natural language feedback within a single sequence. By refining the base model's native uncertainty signals, this manipulation of the latent space enables reliable, granular detection without the overhead of secondary inference loops. Mechanistic analysis via activation patching and layer-wise probing shows that this rank-invariant behavior restructures pre-existing uncertainty geometry into a linearly separable representation that transfers more reliably than base model representations alone. Using tool-calling as an instantiation of granular hallucinations, we validate the detection and downstream improvements enabled by the Latent Critic architecture across Qwen and Llama-based models. Demonstrating superior real-time efficacy, our approach significantly outperforms equivalent-scale fine-tuned external detectors, semantic entropy baselines, and passive internal probes in isolating hallucinations, achieving 0.966 AUROC and $>$80\% accuracy in localization (e.g., \textit{ungrounded: date}). When deployed in a closed-loop ReAct environment, the Critic acts as a negligible latency guardrail, intercepting hallucinations before execution to prevent undesired actions while simultaneously leveraging this specific localized feedback to enable efficient agent self-correction.
\end{abstract}
\section{Introduction}

Human communication is inherently characterized by omission and ambiguity. When an AI agent encounters an incomplete instruction, the optimal behavior is to halt execution, express uncertainty, and seek clarification \citep{vijayvargiya2026ambigsweinteractiveagentsovercome, kim2024aligninglanguagemodelsexplicitly}. However, Large Language Models (LLMs) are heavily optimized for instruction following and next-token prediction which instill a severe task-completion bias \citep{bai2022traininghelpfulharmlessassistant}. In agentic workflows, this bias manifests as a critical failure mode in the form of user specification-grounding failures \citep{sun2025llmshallucinateconnectingdots}. Unlike standard factual hallucinations, an action may be logical, but remains a hallucination if it was never explicitly or implicitly requested by the user. Rather than flagging missing specifications, models confidently execute these undesired actions, creating concrete, misaligned changes to the environment. This lack of adaptability to ambiguous input limits the reliability of AI agents for deployment in high-stakes settings.

Current detection paradigms fail to satisfy the latency, actionability, and generalization requirements of real-time agents. Fine-tuned external LLM-as-a-judge evaluators \citep{info16070517} and multiple-sampling techniques like semantic entropy \citep{ji-etal-2023-towards, kossen2024semanticentropyprobesrobust} operate on surface text, introducing prohibitive inference delays. Furthermore, because models confidently generate these actions, uncertainty-based sampling methods often yield low entropy for these errors, causing detection to collapse. In contrast, probing techniques show that a model's internal representations encode signals of its own uncertainty \citep{orgad2025llmsknowshowintrinsic}. However, passive probing typically trains layer-wise classifiers that yield scalar confidence scores, which lack the actionable localization required to debug an action.

\begin{wrapfigure}{r}{0.4\textwidth}
 \vspace{-10pt}
 \centering
 \includegraphics[width=\linewidth]{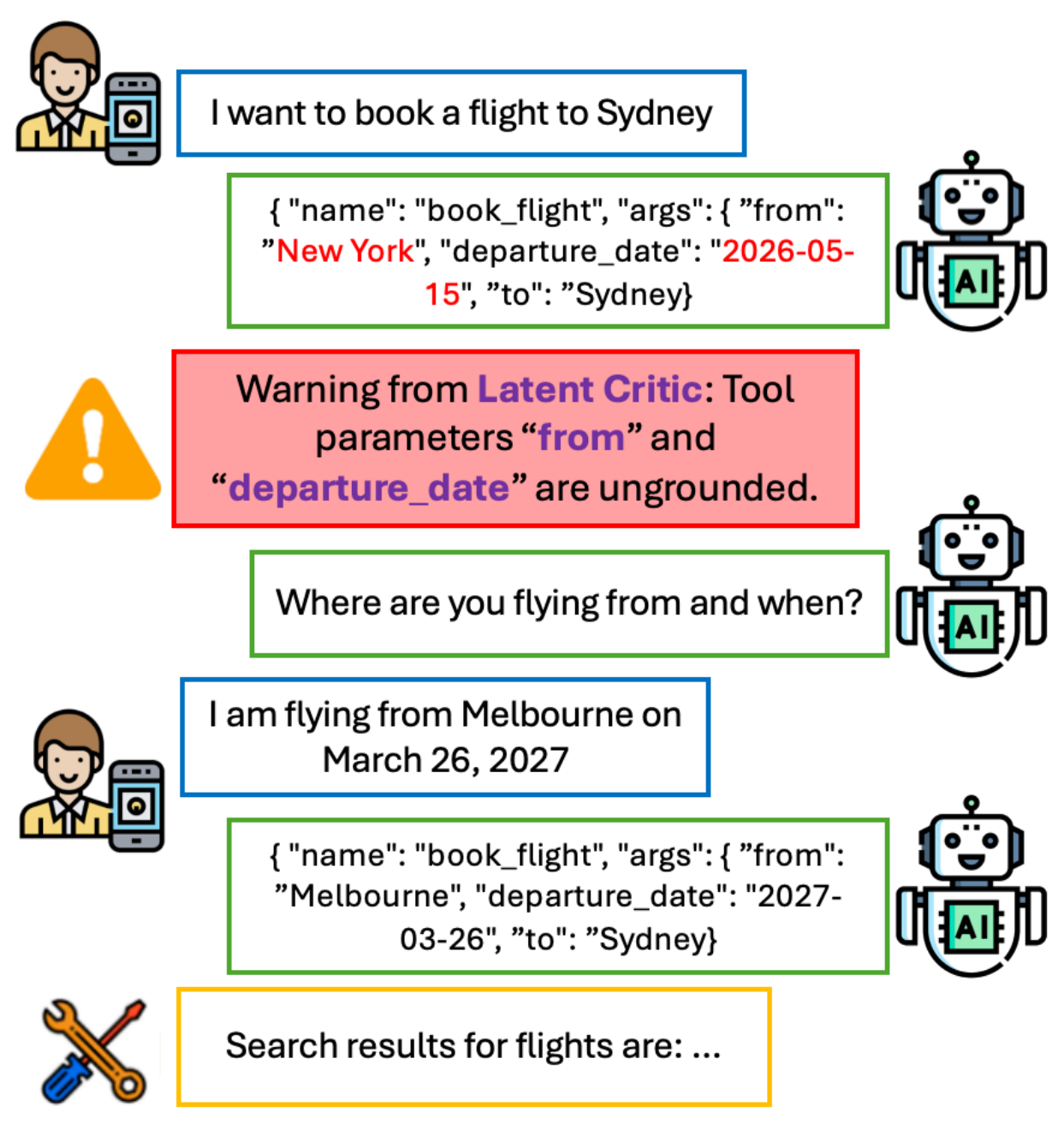}
 \caption{The closed-loop agentic architecture. When the frozen base model generates a hallucinated parameter, the Latent Critic concurrently extracts the specification-grounding failure, blocking execution and feeding the localized error back as a zero-shot environmental observation.}
 \label{fig:agent_loop}
 \vspace{-20pt}
\end{wrapfigure}

This actionability gap raises a fundamental question: \textit{can we translate internal uncertainty into localized feedback without adding prohibitive latency?}

To answer this, we introduce the \textit{Latent Critic}, a Parameter-Efficient Fine-Tuning (PEFT) architecture designed to operate as a minimal-latency mechanistic guardrail. We scope our design and investigation specifically to tool-calling. Unlike open-ended text generation, where the boundaries of a hallucination are often diffuse and subjective, tool calls are executable actions that require exact, parameter-level specification grounding. The structured domain of tool calling acts as a testbed for isolating grounding errors and evaluating fine-grained uncertainty extraction. By attaching a specialized LoRA adapter to a frozen base model, we preserve its native agentic capabilities while leveraging the LLM backbone to verbalize the exact localized hallucination (e.g., \texttt{ungrounded: date}) within the same generation sequence (\autoref{fig:agent_loop}). Through a masked diagnostic objective, the Critic restructures the native representation of specification grounding, amplifying it into a linearly separable geometry that transfers more reliably across distribution shifts than passive probes.

By embedding this detection mechanism directly into the generation process, the Latent Critic bridges the gap between mechanistic representation and practical agent safety. Evaluating this architecture across Qwen and Llama-based models, we demonstrate that exposing and restructuring this latent geometry provides a highly reliable, minimal-latency guardrail that remains the strongest detector across distribution shifts and enables efficient agent self-correction. We structure our investigation around three core Research Questions (RQs):

\begin{itemize}[leftmargin=*, itemsep=0pt]
\item \textbf{RQ1 (Detection):} Can modifying the residual stream yield superior hallucination localization? 
\item \textbf{RQ2 (Mechanism):} How does the adapter restructure the model's internal geometry to translate uncertainty into actionable feedback? 
\item \textbf{RQ3 (Deployment):} Can agents leverage this specific, localized feedback for zero-shot self-correction, easing the safety-productivity tradeoff?
\end{itemize}

\vspace{-5pt}
\section{Experimental Setup}

To rigorously evaluate real-time hallucination detection, we isolate specification-grounding failures from general capability errors and establish a method for generating and labeling tool-calling trajectories at scale.

\subsection{Defining Tool-Calling Hallucinations}
\label{sec:definitions}
Within an agentic tool-calling scope, we evaluate three primary behavioral outcomes: \\
\textbf{(1) Correct \& Grounded (\texttt{ok}):} The correct tool is invoked with all parameters grounded in the contextual history. \\
\textbf{(2) Wrong Tool (\texttt{wrong\_tool}):} A tool unsuitable for the user's goals is invoked (a policy error). \\
\textbf{(3) Ungrounded Parameters (\texttt{ungrounded}):} The correct tool is invoked, but the model fabricates parameters never specified in the context (a specification-grounding error). 

We explicitly exclude capability limitations such as omitted parameters (which occur in $<$0.2\% of our empirical rollouts) and syntax errors. To illustrate the latter: if a user requests an action for "tomorrow" and the agent generates the literal string \texttt{"tomorrow"} instead of the required schema format (e.g., \texttt{"YYYY-MM-DD"}), the model possesses the correct contextual grounding but lacks structural compliance. 

Crucially, our definition of hallucination in this domain focuses entirely on a lack of support in the user's conversational context. Unlike factual hallucinations, which are evaluated against an external ground truth, specification grounding is evaluated strictly against the dialogue history. A generated parameter can be highly plausible or even factually correct (e.g., guessing a valid date or a commonly used setting), but it remains a hallucination if the user never provided it, explicitly or implicitly. 

\subsection{Trajectory Generation}
To generate data characterized by naturalistic ambiguity, we construct a dynamic environment consisting of a base agent and a simulated user \citep{patil2025the,suri2026structureduncertaintyguidedclarification}.

\textbf{Datasets:} Many tool-calling benchmarks feature either simplified schemas that fail to elicit hallucinations at scale, or multi-hop tasks that are prohibitively complex for the evaluated models. To reliably trigger hallucinations, we select base scenarios from SLM agent literature \citep{vijayvargiya2025efficientondeviceagentsadaptive}, providing tractable user intents with highly detailed tool schemas. We utilize 5,000 scenarios for training and a set of 500 tasks with unseen tools for evaluation (ID). We also utilize 200 scenarios from ToolAlpaca \citep{tang2023toolalpacageneralizedtoollearning}---which consists of real-world APIs spanning 50 diverse categories---to evaluate generalization to out-of-distribution (OOD) tasks. The generated tool calls consist of values ranging from single tokens to multiple sentences to evaluate the robustness of the Critic, while the key-value structure allows for precise localization label generation.

\textbf{Simulated User \& Binary Mask:} Powered by Qwen3.5-122B \citep{qwen3.5_2026}, the user incrementally reveals an underlying goal through structurally varied, ambiguous requests. To programmatically track this, the user maintains a hidden \textit{Completed Specifications List (CSL)}---a binary mask over the ground truth tool calls that tracks whether the required information for a parameter in the GT tool calls has been provided (\texttt{1}) or withheld (\texttt{0}).

To illustrate, consider a GT target of \texttt{\{"name": "book\_flight", "args": \{"from": "Melbourne", "to": "Sydney", "date": "2027-03-26"\}\}}. If the user's first turn is "I want to fly to Sydney," the CSL mask initializes as \texttt{\{"from": 0, "to": 1, "date": 0\}}. If the agent then asks for the departure city and the user replies "Melbourne," the mask updates in Turn 2 to \texttt{\{"from": 1, "to": 1, "date": 0\}}.

During the agent's turn, if it outputs a tool call with a parameter corresponding to a \texttt{0}, the generation is labeled as an \texttt{ungrounded} hallucination. Tools entirely absent from the underlying goal are labeled \texttt{wrong\_tool}. The simulated user updates the CSL incrementally at each turn, and the initial programmatic labels with the Qwen 3.5 122B model achieve an 89\% agreement rate with manual human evaluation. As our setting studies specification grounding---whether information is present in the dialogue---rather than conversational naturalness, simulated users provide controlled, balanced interactions that are logistically prohibitive to collect at scale with human subjects \citep{vijayvargiya2026ambigsweinteractiveagentsovercome, suri2026structureduncertaintyguidedclarification}. To ensure rigorous benchmarking, all evaluation trajectories (both ID and OOD) undergo verification by a human annotator, and all reported metrics utilize these corrected, human-verified labels. Failure cases in the programmatic pipeline predominantly included errors in implicit specification tracking or formatting by the simulator model. This pipeline circumvents expensive human labeling for the training set, scales reliably, and accurately localizes exact hallucinated parameters. 

Using our pipeline, we execute rollouts with Qwen3-4B and Llama-xLAM-2-8B \citep{zhang2024xlamfamilylargeaction} to ensure our paradigm generalizes across distinct architectures.

\vspace{-5pt}
\section{RQ1: Actionable Hallucination Detection via the Latent Critic}
\vspace{-5pt}

Existing internal probing techniques yield passive scalar outputs that lack the verbalized localization required for agentic self-correction. To address RQ1, we introduce the Latent Critic designed to concurrently extract internal uncertainty and output the precise semantic location of a hallucination (e.g., specific keys).

\vspace{-5pt}
\subsection{Latent Critic Architecture}
\begin{figure}[t]
 \centering
 \includegraphics[width=0.9\textwidth]{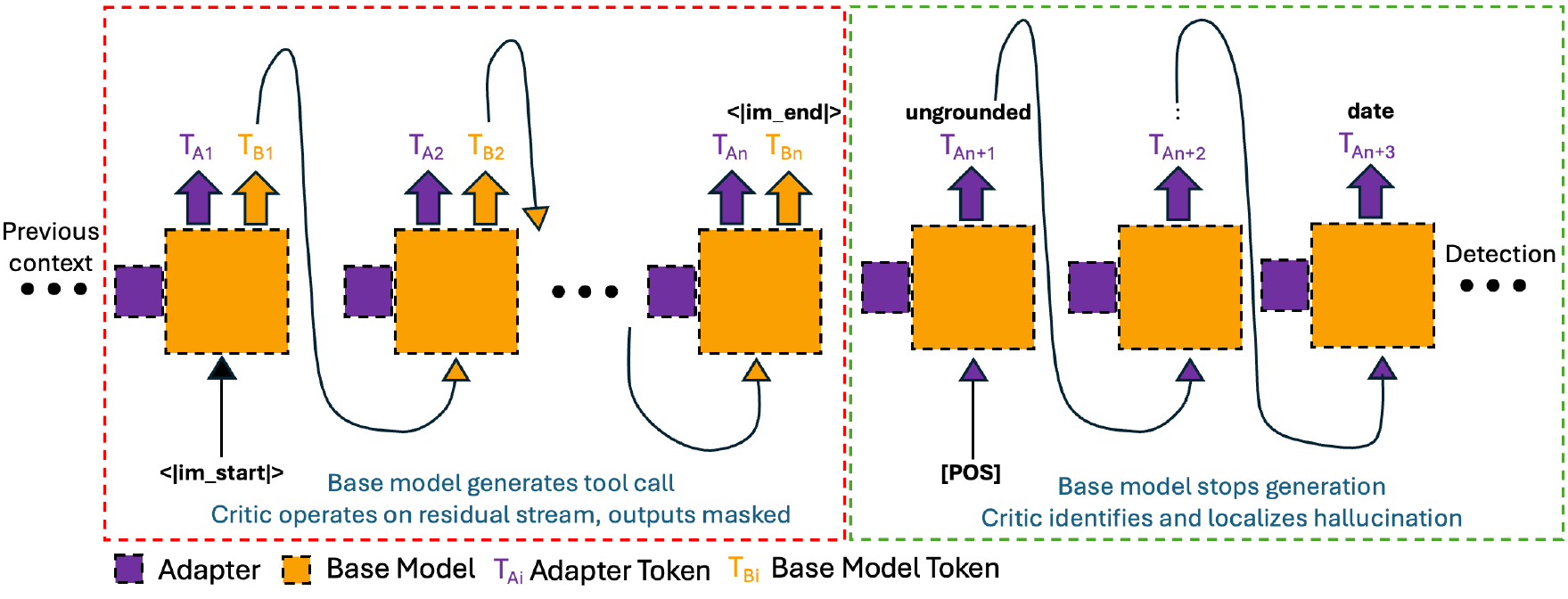}
 \caption{Concurrent extraction architecture. The adapter processes the base model's hidden states during tool call generation and projects the accumulated uncertainty upon generating \texttt{[POS]} token.}
 \label{fig:rq1_arch}
 \vspace{-10pt}
\end{figure}

\vspace{-5pt}
The detection module consists of a Low-Rank Adaptation (LoRA) module~\citep{hu2021loralowrankadaptationlarge} executed concurrently with the frozen base agent via multi-adapter serving. Because the adapter's weights are initialized at zero, its initial forward pass mirrors the base model's exact hidden states. During the agentic loop, the frozen base model autoregressively generates the tool call syntax. Concurrently, the adapter receives the same input tokens but learns to modify hidden states and amplify uncertainty signals, while its outputs are disregarded in this phase (\autoref{fig:rq1_arch}). 

Upon completion of the tool call, a trigger token (\texttt{[POS]}) is appended, acting as a localized attention sink. This token establishes a structural boundary, signaling that the Critic must now verbalize the detection results by aggregating uncertainty from the full tool-call context and generating a structured classification (e.g., \texttt{ok}, \texttt{wrong\_tool}, or \texttt{ungrounded: [parameter\_name]}). To ensure the adapter acts strictly as an uncertainty extractor, we modify the Supervised Fine-Tuning (SFT) objective. During loss calculation, the base tool-call generation is masked and the gradients are computed only on the \texttt{[POS]} token and the subsequent label. It is important to note that during inference, the adapter actively processes inputs while the base model generates the tool call, however, its outputs at this stage are disregarded. This draws inspiration from pause tokens~\citep{goyal2024thinkspeaktraininglanguage}. Yet, while prior work utilizes dummy tokens to inject extra computational steps during text generation, the Critic's steps during the base model's tool call generation allow for uncertainty accumulation.

\begin{table*}[t]
\centering
\resizebox{\textwidth}{!}{
\begin{tabular}{l ccc ccc}
\toprule
\multirow{2}{*}{\textbf{Method}} & \multicolumn{3}{c}{\textbf{In-Distribution (ID)}} & \multicolumn{3}{c}{\textbf{Out-of-Distribution (OOD)}} \\
\cmidrule(lr){2-4} \cmidrule(lr){5-7}
& \textbf{F1} & \textbf{AUROC} & \textbf{AUPRC} & \textbf{F1} & \textbf{AUROC} & \textbf{AUPRC} \\
\midrule
Token Entropy & 0.668 $\pm$ 0.031 & 0.674 $\pm$ 0.035 & 0.626 $\pm$ 0.050 & 0.370 $\pm$ 0.032 & 0.485 $\pm$ 0.039 & 0.222 $\pm$ 0.031 \\
Semantic Entropy & 0.623 $\pm$ 0.031 & 0.639 $\pm$ 0.036 & 0.619 $\pm$ 0.049 & 0.369 $\pm$ 0.032 & 0.460 $\pm$ 0.041 & 0.220 $\pm$ 0.032 \\
SEP (Kossen et al.) & 0.623 $\pm$ 0.031 & 0.548 $\pm$ 0.038 & 0.482 $\pm$ 0.045 & 0.372 $\pm$ 0.038 & 0.499 $\pm$ 0.039 & 0.235 $\pm$ 0.035 \\
Linear Probe & 0.697 $\pm$ 0.031 & 0.782 $\pm$ 0.024 & 0.760 $\pm$ 0.033 & 0.383 $\pm$ 0.027 & 0.616 $\pm$ 0.029 & 0.278 $\pm$ 0.032 \\
External Judge & 0.695 $\pm$ 0.041 & 0.915 $\pm$ 0.016 & 0.864 $\pm$ 0.029 & 0.647 $\pm$ 0.021 & 0.884 $\pm$ 0.020 & 0.741 $\pm$ 0.041 \\
\textbf{Latent Critic (ours)} & \textcolor{blue}{\textbf{0.870 $\pm$ 0.025}} & \textcolor{blue}{\textbf{0.966 $\pm$ 0.010}} & \textcolor{blue}{\textbf{0.924 $\pm$ 0.027}} & \textcolor{blue}{\textbf{0.670 $\pm$ 0.033}} & \textcolor{blue}{\textbf{0.925 $\pm$ 0.015}} & \textcolor{blue}{\textbf{0.815 $\pm$ 0.034}} \\
\bottomrule
\end{tabular}
}
\caption{Hallucination Detection Performance on \textbf{Qwen3-4B}. Positive class = \texttt{ungrounded}. Critic and Judge evaluated one-vs-rest on the full 3-way set (ID N=303: \texttt{ok} 41.6\%, \texttt{ug} 35.0\%, \texttt{wt} 23.4\%; OOD N=522: \texttt{ok} 76.2\%, \texttt{ug} 20.7\%, \texttt{wt} 3.1\%). Probe and Entropy family evaluated strictly on the binary \texttt{ok}-vs-\texttt{ungrounded} subset. $\pm$ denotes one standard deviation over 10,000 bootstrap resamples. The Latent Critic achieves peak performance across all threshold-independent metrics.}
\label{tab:qwen_results}
\vspace{-10pt}
\end{table*}

\begin{table}[t]
\centering
\begin{tabular}{l cc}
\toprule
\multirow{2}{*}{\textbf{Method}} & \multicolumn{2}{c}{\textbf{Llama-xLAM-2-8B}} \\
\cmidrule(lr){2-3}
& \textbf{ID AUROC} & \textbf{OOD AUROC} \\
\midrule
Linear Probe & 0.833 $\pm$ 0.044 & 0.514 $\pm$ 0.030 \\
External Judge & 0.903 $\pm$ 0.031 & 0.630 $\pm$ 0.030 \\
\textbf{Latent Critic (ours)} & \textcolor{teal}{\textbf{0.926 $\pm$ 0.029}} & \textcolor{teal}{\textbf{0.649 $\pm$ 0.028}} \\
\bottomrule
\end{tabular}
\vspace{5pt}
\caption{Cross-Model Replication on \textbf{Llama-xLAM-2-8B}. Positive class = \texttt{ungrounded}. The advantage over passive-probing strongly replicates, while the advantage over external evaluators proves model-dependent under shift.}
\label{tab:llama_results}
\vspace{-15pt}
\end{table}

\vspace{-5pt}
\subsection{Results}
\vspace{-5pt}
We evaluate the Critic across Qwen3-4B and Llama-xLAM-2-8B against an equivalent-scale External Judge (Qwen 3.5 4B), a passive internal Linear Probe trained on the base model's optimal layer, and multiple uncertainty-based baselines (Token Entropy, Semantic Entropy, and SEP). The External Judge and the Linear Probe are trained with the same data as the Critic, with the Judge using the exact configuration as the Critic (LoRA rank, input prompt, labels, etc.) while the probe and entropy methods are tuned on the ok vs ungrounded classification only (no wrong\_tool). We hypothesize that actively restructuring the residual stream allows for granular, robust detection that external text-based evaluators, entropy methods, and passive internal probes miss. Detailed class-wise Precision, Recall, and F1 metrics are available in Appendix \ref{app:full_metrics}.

\textbf{In-Distribution Detection:} Evaluated on our test set, the Latent Critic demonstrates the strongest overall performance (Table \ref{tab:qwen_results}). Crucially, these results highlight the fundamental nature of specification-grounding errors. Because models confidently fabricate plausible parameters to complete tasks, standard uncertainty-based sampling methods yield low entropy for these errors. Consequently, methods like Semantic Entropy collapse into a single cluster; when adjusting their AUPRC metrics for baseline prevalence, they sit exactly at the trivial always-positive decision floor, failing to discriminate grounding entirely. 

Forced to judge surface-level text, the External Judge struggles with calibration. While it ranks well on Qwen ($0.915 \pm 0.016$ AUROC), it generates poorly (0.695 F1). By directly modifying the hidden states, the Latent Critic isolates the actual geometric signal of specification grounding, achieving 0.966 AUROC on Qwen and localizing errors with $>$80\% exact parameter-match accuracy. Importantly for deployment where minimizing false blocks is paramount, the Critic successfully catches 38.7\% of Qwen's hallucinations at a strict 1\% False Positive Rate (FPR), compared to 28.3\% for the Judge and just 3.8\% for the passive Probe.

\textbf{Out-of-Distribution Generalization:} To test robustness under shift, we evaluate these detectors on unseen scenarios from ToolAlpaca (Table \ref{tab:qwen_results}). The Linear Probe collapses ($0.782 \rightarrow 0.616$ AUROC), demonstrating that un-restructured internal representations of grounding do not transfer reliably. In contrast, the Latent Critic maintains a strong $0.925 \pm 0.015$ AUROC. The drop in its OOD F1 score ($0.870 \rightarrow 0.670$) is partly driven by base-rate mechanics, as the OOD prevalence of hallucinations drops from 35.0\% to 20.7\%. The entropy baselines remain non-discriminative under shift.

\textbf{Cross-Model Replication:} We also validate this architecture on Llama-xLAM-2-8B (Table \ref{tab:llama_results}). The advantage of restructuring over passive probing strongly replicates across both settings (Critic 0.926 AUROC vs. Probe 0.833 ID; Critic 0.649 vs. Probe 0.514 OOD). However, the advantage of internal access over the external judge under shift appears model-dependent. On the xLAM backbone, both the Critic and Judge degrade similarly (0.649 vs. 0.630 AUROC). Ultimately, however, the Latent Critic's improvement over passive internal probing remains statistically significant across both models ($p<0.001$, measured via McNemar's test~\citep{McNemar_1947}).

\textbf{Error Analysis:} While the metrics represent a significant improvement over existing methods, the Latent Critic's accuracy remains fundamentally bottlenecked by the base model's own capability failures. Because it reads internal states, the Critic is reliable only if the base model internally registers the specification gap. Consequently, detection errors predominantly stem from base model breakdowns that corrupt this geometric signal. For instance, during runaway generation---where the model infinitely loops---the residual stream becomes chaotic, destroying the uncertainty geometry. Similarly, partial grounding dilutes the signal: if a user specifies "2 PM" but the schema requires a full \texttt{datetime}, the agent anchors on the grounded time while confidently fabricating the date, occasionally bypassing the Critic.

Perceived errors also reveal the Critic's nuanced grounding boundaries. If instructed to "Send an email about X," an agent might create a plausible subject and body. While human judges might accept this, the Critic strictly flags it as \texttt{ungrounded}, penalizing unguided, high-stakes hallucination. Yet, for tasks inherently requiring generative expansion (e.g., text summarization), it correctly permits the output. Finally, occasional conflation with the \texttt{wrong\_tool} class occurs because policy/routing errors can overlap with grounding signals. Overall, misclassifications usually stem from the base model's unreliability in generation rather than extraction flaws, hinting the Critic's real-world utility exceeds its strict metrics.

\vspace{-3pt}
\section{RQ2: Mechanistic Analysis of Latent Uncertainty Extraction}
\vspace{-2pt}
While RQ1 establishes the empirical strengths of the Latent Critic, RQ2 investigates the internal mechanisms enabling this behavior using Qwen3-4B. We analyze how our adapter architecture geometrically maps the uncertainty in the residual stream to the vocabulary space. Specifically, we structure our analysis to establish \textit{why} internal access is necessary (Section~\ref{subsec:why}), \textit{when} the specification-grounding signal emerges (Section~\ref{subsec:when}), \textit{what} the adapter does to these representations (Section~\ref{subsec:what}), and provide \textit{causal evidence} that the adapter relies on this underlying geometry (Section~\ref{subsec:causal}).

\begin{figure*}[t]
 \centering
 \begin{subfigure}[b]{0.45\textwidth}
  \centering
  \includegraphics[width=\textwidth]{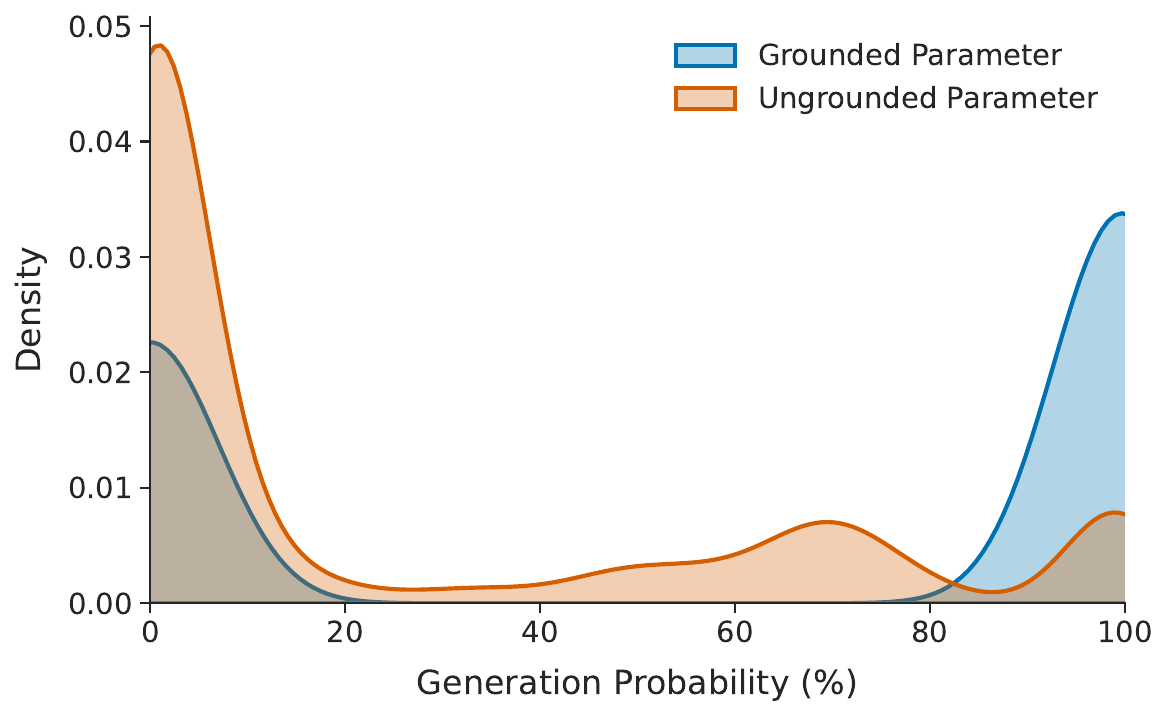}
  \caption{\textbf{Post-Softmax Confidence.} Distribution overlap renders simple logit-thresholding nonviable.}
  \label{fig:mech_density}
 \end{subfigure}
 \hfill
 \begin{subfigure}[b]{0.45\textwidth}
  \centering
  \includegraphics[width=\textwidth]{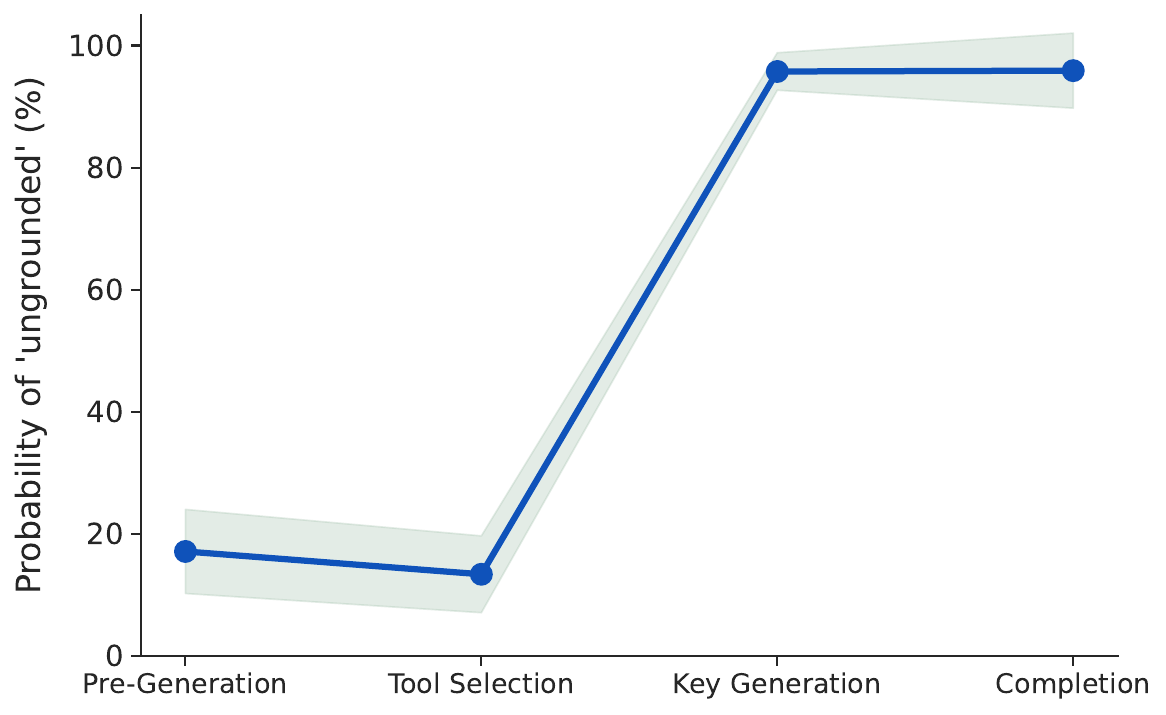}
  \caption{\textbf{Evolution of Uncertainty.} Sharp spike to $>$95\% when the ungrounded key is formulated.}
  \label{fig:mech_timing}
 \end{subfigure}
  
 \vspace{10pt} 
  
 \begin{subfigure}[b]{0.45\textwidth}
  \centering
  \includegraphics[width=\textwidth]{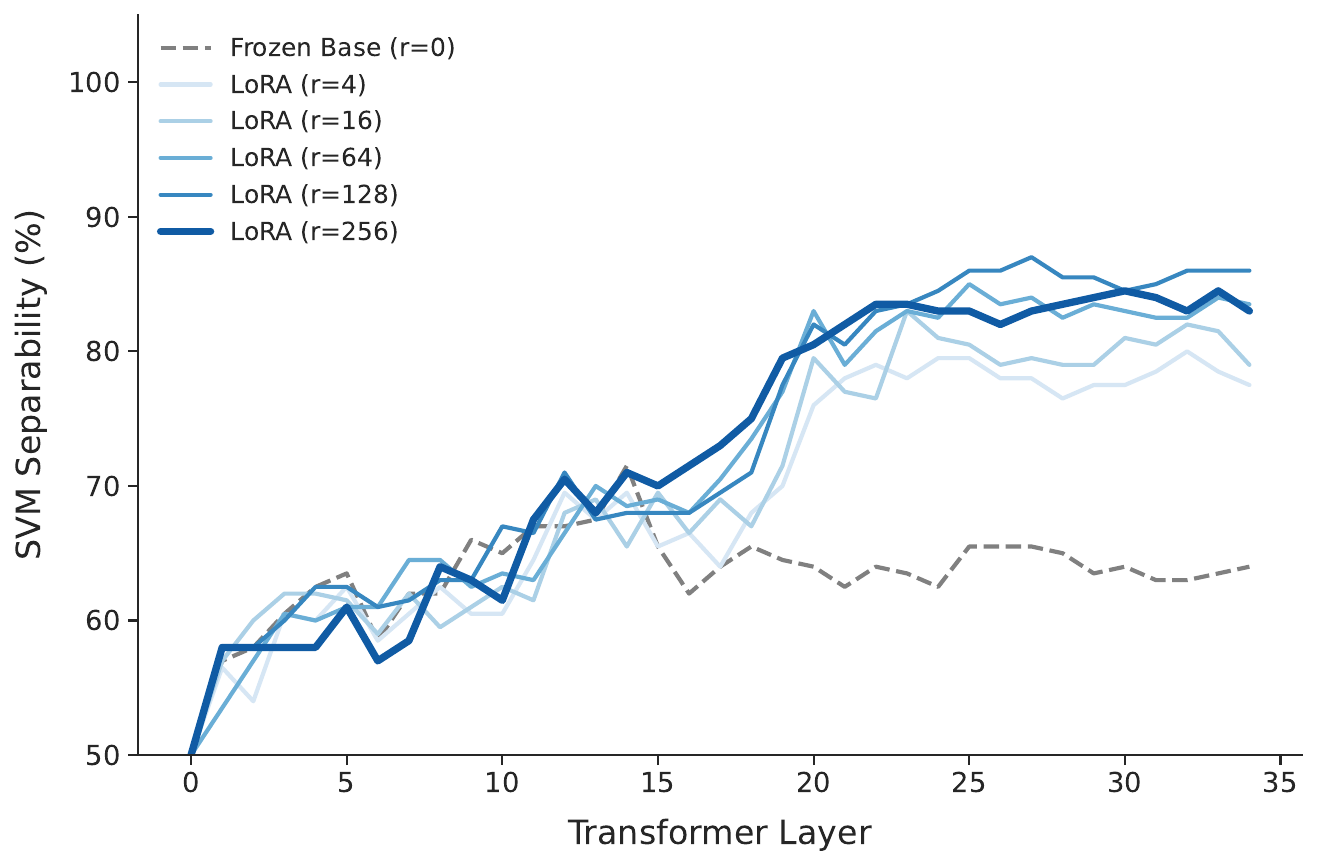}
  \caption{\textbf{Linear Separability.} Latent Critic states undergo sharp, rank-invariant disentanglement around Layer 15.}
  \label{fig:mech_svm}
 \end{subfigure}
 \hfill
 \begin{subfigure}[b]{0.45\textwidth}
  \centering
  \includegraphics[width=\textwidth]{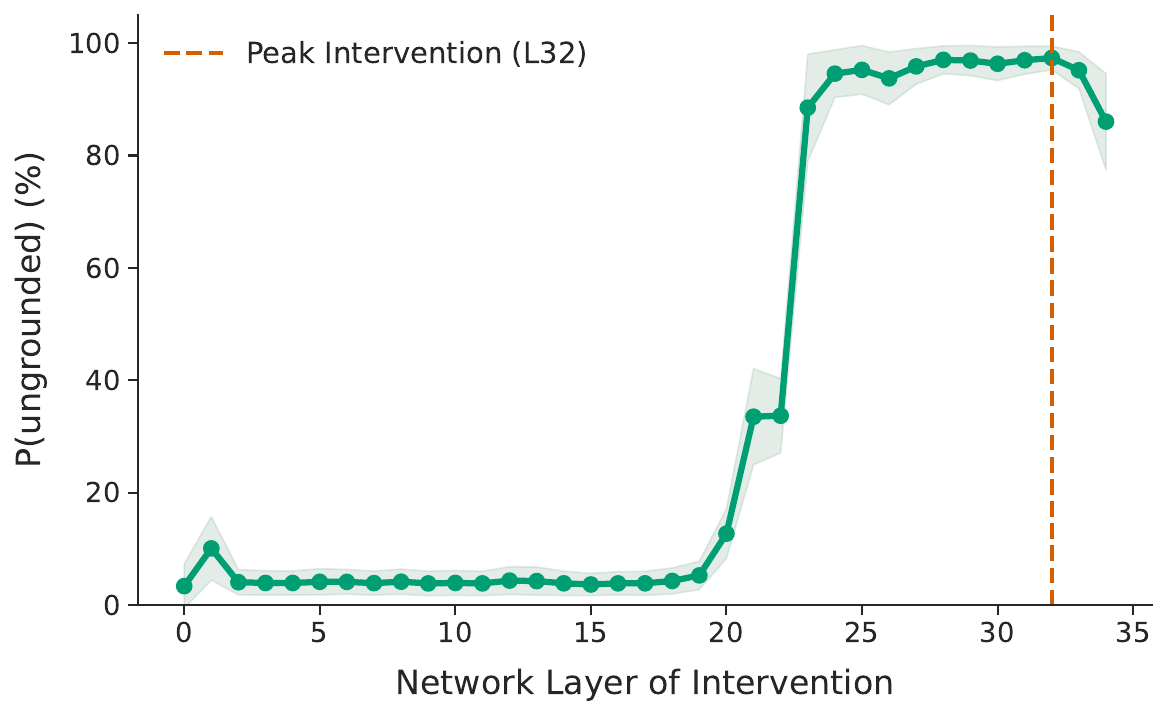}
  \caption{\textbf{Causal Tracing.} Transplanting a corrupted state in late layers overwrites the network's prediction.}
  \label{fig:mech_patching}
 \end{subfigure}
 \vspace{-3pt}
 \caption{\textbf{Mechanistic Analysis of the Latent Critic.} (a) Surface-level output logits fail to reliably separate grounded parameters from fabrications. (b) Latent extraction is reactive, not prescient. (c) The adapter acts as a geometric amplifier, separating entangled base states. (d) Localized activation patching confirms the adapter relies causally on this underlying latent geometry.}
 \label{fig:mechanisms}
 \vspace{-15pt}
\end{figure*}

\vspace{-5pt}
\subsection{Confidence in Predictions Does Not Imply Reliability}
\label{subsec:why}
To establish \textit{why} external judges and entropy methods struggle, we analyze the token-generation probabilities of the base model across 500 scenarios. Standard logit-based confidence scoring fails because models are optimized for task completion. They suffer from poor confidence calibration \citep{azaria2023internalstatellmknows}, regularly outputting confident fabrications (e.g., guessing plausible tokens with $>$80\% confidence) while also generating low-confidence grounded actions. Consequently, the probability density functions of grounded and ungrounded parameters exhibit a strong overlap (Figure \ref{fig:mechanisms}a), making simple logit-thresholding unreliable. By processing the hidden states directly, the Latent Critic circumvents this constraint. It classifies the ungrounded trajectory with an average confidence of \textbf{95.38\%}, demonstrating that internal geometry captures significantly more grounding information than is permitted to surface in output logits \citep{kossen2024semanticentropyprobesrobust}.

\vspace{-5pt}
\subsection{Latent Extraction is Reactive, Not Prescient}
\label{subsec:when}
\vspace{-5pt}
To determine exactly \textit{when} the specification-grounding signal arises, we track the Latent Critic's classification confidence across four distinct stages of tool-call construction ($N=500$). We incrementally append the \texttt{[POS]} trigger token: (1) before the tool call, (2) after the tool name was selected, (3) immediately after the ungrounded parameter key was generated, and (4) upon completion of the full payload.

As shown in Figure \ref{fig:mechanisms}b, the average probability of outputting \texttt{ungrounded} remains volatile prior to generating the payload (\textbf{19.54\%}) and drops slightly when the correct tool name is selected (\textbf{16.25\%}), as tool selection acts as a coarse-grained commitment that partially resolves the trajectory. However, at the exact sequence step where the ungrounded parameter key is generated, confidence sharply spikes to \textbf{95.24\%}. This establishes that the internal representation is reactive rather than predictive: the signal denoting a hallucination emerges within the hidden states the instant the model formulates the flawed parameter. 

\begin{table}[t]
\small
\centering
\begin{tabular}{lcc}
\toprule
\textbf{Probe Configuration} & \textbf{ID AUROC} & \textbf{OOD AUROC} \\
\midrule
Linear Probe (Frozen Base States) & 0.782 & 0.616 \\
2-Layer MLP (Frozen Base States) & 0.808 & 0.678 \\
Linear Probe (Latent Critic States) & 0.944 & 0.915 \\
\textbf{Latent Critic (Full Output)} & \textbf{0.966} & \textbf{0.925} \\
\bottomrule
\end{tabular}
\vspace{5pt}
\caption{Representation Transfer Analysis. Probes were trained on 16k examples using states extracted from the final generated tool-call token at Layer 21. Training a linear probe on the Latent Critic's restructured states bridges the OOD generalization gap, confirming the adapter exposes a shift-stable geometry rather than simply memorizing a classifier.}
\label{tab:probe_ladder}
\vspace{-15pt}
\end{table}

\vspace{-5pt}
\subsection{Rank-Invariant Restructuring of the Residual Stream}
\label{subsec:what}
\vspace{-5pt}
To understand \textit{what} the Critic does to these representations, we analyze the linear separability of the hidden states. As established in RQ1, the frozen base model's raw hidden states contain a partial linear representation of grounding In-Distribution, but this direction fails to transfer under shift (0.616 AUROC). 

To determine whether this transfer failure is due to limited probe capacity or the underlying representation itself, we trained a 2-layer MLP probe on the frozen states (Table \ref{tab:probe_ladder}). The increased capacity yielded only a modest OOD improvement (0.678 AUROC). However, training a simple linear probe on the \textit{Latent Critic's} adapted hidden states achieved 0.944 ID and 0.915 OOD AUROC. Because the linear probe becomes highly effective once trained on the adapted states, it provides strong evidence that the masked objective explicitly restructures the representation into a shift-stable linear geometry, rather than the adapter memorizing a complex, non-linear classifier from scratch.

Furthermore, as illustrated in Figure \ref{fig:mechanisms}c, this geometric transformation is rank-invariant. Although indistinguishable from the base model in early syntactic layers (0–14), the adapted representations undergo a large increase in separability starting at Layer 15. Even highly bottlenecked adapters (e.g., $r=4$) successfully separate the latent space. This implies that the specification-grounding signal occupies a low-dimensional subspace; the adapter only needs to learn a low-rank linear projection to isolate and amplify it.

\subsection{Causal Verification via Activation Patching}
\label{subsec:causal}
Finally, to obtain \textit{causal evidence} that the adapter operates strictly on latent geometric representations rather than surface-level text, we perform cross-trajectory activation patching \citep{meng2023locatingeditingfactualassociations}. Restricting our intervention to the final \texttt{[POS]} token, we execute a forward pass on a corrupted, ungrounded trajectory and cache the hidden states. We then execute a second forward pass on a grounded trajectory. During this second pass, we overwrite the clean \texttt{[POS]} token's hidden state at layer $L$ with the corrupted state, allowing the network to complete the remaining forward pass naturally. The context window and prior generated tokens remain grounded. Unlike standard activation patching, which corrupts a single input, we patch across different trajectories to test whether the geometric signal generalizes across contexts.

As shown in Figure \ref{fig:mechanisms}d, despite the surrounding text indicating a correct tool-call, this localized intervention forces the Latent Critic to output \texttt{ungrounded} in \textbf{96\%} of trajectories. Executing this layer-by-layer reveals a distinct S-curve. Patching early syntactic layers (0–20) yields negligible changes ($<$5\% probability), while patching mid-to-late layers results in a sharp probability inversion peaking at Layer 32. This provides strong causal evidence that the hallucination signal is a geometric feature formed in the later layers, and that the adapter's classification operates independently of surface text. Together, this analysis provides an understanding of how the Latent Critic performs its function, the role of the trigger token, and why it is effective.
\vspace{-5pt}
\section{RQ3: Closed-Loop Agentic Self-Correction}

Having established that the Latent Critic accurately extracts parameter-specific errors, we evaluate its utility within an agentic loop. We hypothesize that while simply blocking an action provides safety, projecting that latent uncertainty into targeted natural language (e.g., \texttt{ungrounded: date}) functions as an actionable environmental observation, enabling the base agent to self-correct more efficiently than a generic safety block.

\vspace{-5pt}
\subsection{Experimental Setup}
To test the real-world applicability of our architecture, we integrate the Latent Critic into a closed-loop ReAct environment. We evaluate 500 in-distribution (ID) trajectories derived from \cite{vijayvargiya2025efficientondeviceagentsadaptive}, alongside 200 out-of-distribution (OOD) trajectories featuring unseen API schemas and scenarios sampled from ToolAlpaca \citep{tang2023toolalpacageneralizedtoollearning}. We compare the base agent loop against two intervention paradigms to isolate the utility of our generative architecture. 
\vspace{-3pt}
\begin{enumerate}[leftmargin=*, itemsep=0pt]
  \item \textbf{Generic Intervention:} Simulates a standard binary classification probe by using the Latent Critic for detection without providing localized feedback. If an error is detected, the environment blocks the tool and returns a generic error: \textit{System Error: Tool execution blocked due to hallucination.} This isolates the marginal utility of the Critic's verbalization.
  \item \textbf{Specific Intervention:} Our proposed method. The environment blocks the tool and injects the specific, localized verbalization: \textit{System Error: Tool blocked. You hallucinated the parameter: '[PARAM]'.}
\end{enumerate}
\vspace{-3pt}

To rigorously evaluate the impact of the Latent Critic, we isolate the model's tool-calling capabilities by measuring \textit{Parameter-Level Metrics} across both ID and OOD datasets. Rather than judging a multi-turn trajectory solely as a binary success or failure, we evaluate the exact overlap between the agent's executed tool call arguments and the ground-truth parameters required to fulfill the user's intent. In this formulation, \textbf{Precision} acts as a strict proxy for safety: out of all the parameters the agent successfully executed, what percentage were actually grounded and correct? Conversely, \textbf{Recall} measures functional task completion: out of all the parameters required by the user's goal, what percentage did the agent successfully execute? The \textbf{F1-Score} provides the harmonic mean, representing the agent's overall functional reliability. We also report \textbf{Trajectory Success} and the \textbf{False Block Rate} (the percentage of correctly grounded calls erroneously blocked). We only consider successfully executed tool calls for parameter analysis; blocked calls are disregarded.

\vspace{-5pt}
\subsection{Results}
\vspace{-5pt}
\begin{table}[htbp]
\small
\vspace{-5pt}
\centering
\resizebox{\columnwidth}{!}{
\begin{tabular}{l ccc | cc}
\toprule
& \multicolumn{3}{c}{\textbf{Parameter-Level Metrics (\%)}} & \multicolumn{2}{c}{\textbf{Trajectory-Level (\%)}} \\
\cmidrule(lr){2-4} \cmidrule(lr){5-6}
\textbf{Agent Configuration} & \textbf{Precision} & \textbf{Recall} & \textbf{F1-Score} & \textbf{Success Rate} & \textbf{False Block} \\
\midrule
\multicolumn{6}{c}{\textit{In-Distribution (ID)}} \\
\midrule
Base Agent (No Intervention) & 49.7 $\pm$ 2.2 & \textbf{54.8 $\pm$ 2.6} & 52.1 $\pm$ 1.9 & 16.4 $\pm$ 1.8 & -- \\
+ Generic Intervention & 66.1 $\pm$ 3.4 & 41.9 $\pm$ 2.4 & 51.3 $\pm$ 2.2 & 20.3 $\pm$ 1.8 & 9.4 $\pm$ 2.2 \\
\textbf{+ Specific (Ours)} & \textbf{69.8 $\pm$ 1.8} & 54.4 $\pm$ 2.2 & \textbf{61.2 $\pm$ 1.8} & \textbf{22.1 $\pm$ 1.9} & \textbf{2.9 $\pm$ 1.0} \\
\midrule
\multicolumn{6}{c}{\textit{Out-of-Distribution (OOD)}} \\
\midrule
Base Agent (No Intervention) & 52.5 $\pm$ 3.4 & \textbf{26.6 $\pm$ 2.1} & 35.3 $\pm$ 2.2 & \textbf{2.2 $\pm$ 0.8} & -- \\
+ Generic Intervention & 62.3 $\pm$ 4.7 & 21.3 $\pm$ 2.8 & 31.7 $\pm$ 2.4 & 2.0 $\pm$ 1.4 & 15.9 $\pm$ 4.1 \\
\textbf{+ Specific (Ours)} & \textbf{66.8 $\pm$ 3.0} & 25.3 $\pm$ 2.7 & \textbf{36.7 $\pm$ 1.9} & 2.1 $\pm$ 1.5 & \textbf{13.2 $\pm$ 3.3} \\
\bottomrule
\end{tabular}
}
\vspace{5pt}
\caption{Impact of intervention paradigms on agentic reliability. Generic detection boosts Precision (acting as a safety net) but crashes Recall due to a high false block rate and confused retry loops. The Latent Critic's \textit{Specific} verbalization uniquely cushions this tradeoff, minimizing false blocks and maximizing parameter F1.}
\label{tab:rq3_downstream}
\vspace{-25pt}
\end{table}

Because the simulated user deliberately provides ambiguous instructions, the Base Agent exhibits a severe task-completion bias, resulting in ungrounded, low-precision calls attempting to force task completion (\autoref{tab:rq3_downstream}). Applying a \textbf{Generic Intervention} acts as a safety boundary, driving ID precision up to 66.1\% and OOD precision to 62.3\% and blocks these actions. However, because the agent is not given localized feedback, it cannot effectively correct the payload. Due to repeated hallucination blocks, an elevated false-block rate (9.4\%), and a lack of actionable feedback, fewer actions can be executed successfully, resulting in a drop in recall and overall parameter F1.

The \textbf{Specific Latent Critic} effectively mitigates this tradeoff. On ID tasks, it achieves the highest precision (69.8\%) while successfully guiding the agent to recover its recall (54.4\%), yielding a peak parameter F1 score of 61.2\%. Specific feedback concurrently improves Trajectory Success to 22.1\% while reducing false blocks by more than 3$\times$ (down to 2.9\%) compared to generic feedback. 

Under distribution shift (OOD), trajectory success remains close to the frozen base model's ceiling ($\sim$2\%) across all methods, making trajectory completion metrics relatively insensitive due to the base model's broader capability failures on unseen datasets. However, parameter-level recovery still clearly improves (OOD F1: 36.7\% vs 35.3\%; Precision: 66.8\% vs 52.5\%), indicating that the Critic continues to identify the correct interventions and enforce safety even when the base agent ultimately lacks the reasoning capacity to fully complete the task.

\vspace{-8pt}
\subsection{Self-Correction}
\vspace{-5pt}
To quantify how specific verbalization affects execution metrics, we isolate the subset of trajectories where an intervention occurred and track the agent's ability to self-correct its policy (e.g., pivoting to ask the user a clarifying question instead of engaging in repeated guessing).

\begin{table}[htbp]
\vspace{-5pt}
\small
\centering
\begin{tabular}{l cc | cc}
\toprule
& \multicolumn{2}{c}{\textbf{In-Distribution (ID)}} & \multicolumn{2}{c}{\textbf{Out-of-Distribution (OOD)}} \\
\cmidrule(lr){2-3} \cmidrule(lr){4-5}
\textbf{Intervention} & \textbf{Recovery Rate (\%)} & \textbf{Failed Attempts} & \textbf{Recovery Rate (\%)} & \textbf{Failed Attempts} \\
\midrule
Generic & 23.9 $\pm$ 2.7 & 1.47 $\pm$ 0.14 & 20.0 $\pm$ 5.3 & 1.41 $\pm$ 0.12 \\
\textbf{Specific (Ours)} & \textbf{37.0 $\pm$ 1.9} & \textbf{1.06 $\pm$ 0.09} & \textbf{29.3 $\pm$ 4.7} & \textbf{1.00 $\pm$ 0.00} \\
\bottomrule
\end{tabular}
\vspace{5pt}
\caption{\textbf{Self-Recovery Mechanics.} Providing exact parameter localization improves the agent's recovery rate while significantly reducing wasted inference loops across both distributions.}
\label{tab:rq3_recovery}
\vspace{-13pt}
\end{table}

As detailed in Table \ref{tab:rq3_recovery}, actionable specific verbalization improves both the rate of recovery and its efficiency. When faced with a Generic block, the agent often enters a failure loop, repeatedly guessing values for different parameters or trying new hallucinated values for the same parameter (\ref{app:failure_walkthroughs}). By injecting the exact diagnostic string, the Latent Critic boosts the successful recovery rate to \textbf{37.0\%} (ID) and \textbf{29.3\%} (OOD), representing relative improvements of 54.8\% and 46.5\%, respectively. 

Furthermore, it consistently reduces the average number of failed retry attempts. On novel OOD schemas, the Latent Critic achieves an average of only \textbf{1.00 attempt} prior to success. However, it is important to contextualize this efficiency. In many cases, the agent never recovers, signaling a need for dedicated recovery-policy training to fully utilize the diagnostic feedback. Nevertheless, in cases where successful recoveries are possible with the frozen base agent, exact parameter localization significantly reduces the computational overhead and latency associated with retry loops. Finally, as a single-pass extraction module, the Critic itself operates with negligible overhead, adding $<$10 ms per executed tool call in our vLLM deployment environment (and 58 ms in a standalone HuggingFace setup), satisfying the strict real-time requirements of autonomous agents.
\vspace{-5pt}
\section{Related Work}
\vspace{-5pt}
\textbf{Agentic Tool-Calling and Grounding.} Existing evaluation frameworks for AI agents \citep{li2023apibankcomprehensivebenchmarktoolaugmented, qin2023toolllmfacilitatinglargelanguage, patil2025the} and constrained decoding pipelines \citep{beurerkellner2024guidingllmsrightway} predominantly emphasize functional correctness and syntactic validity. However, when faced with underspecified instructions, LLMs succumb to a task-completion bias, fabricating missing parameters rather than seeking clarification \citep{zhang-choi-2025-clarify, chi2024clarinetaugmentinglanguagemodels, vijayvargiya2026askingmattersrewarddrivenclarification}. To mitigate these specification-grounding failures, the dominant paradigm relies on post-hoc action correction \citep{shinn2023reflexionlanguageagentsverbal, gou2024criticlargelanguagemodels}. Yet, because these methods operate on surface text, they incur secondary inference latency and are susceptible to hallucinations. Crucially, they require an erroneous action to be executed and observed to fail before correction begins, whereas our approach intercepts the action pre-execution.

\textbf{Real-Time Hallucination Detection.} Detection methods span from output-level sampling \citep{manakul2023selfcheckgptzeroresourceblackboxhallucination, kuhn2023semanticuncertaintylinguisticinvariances}---which is robust but incurs prohibitive latency for real-time deployment---to single-pass internal probes \citep{kadavath2022languagemodelsmostlyknow, azaria2023internalstatellmknows, su2024unsupervisedrealtimehallucinationdetection, kossen2024semanticentropyprobesrobust}. While probing resolves latency bottlenecks by training lightweight classifiers directly on hidden states, it functions as a passive observer, outputting scalars that are non-actionable for granular self-correction. Furthermore, existing works primarily evaluate factual correctness. In agentic workflows, the challenge shifts to \textit{user-specification alignment} where a guessed parameter (e.g., a valid date) may be factually plausible but remains a hallucination if not grounded in the user's context. While recent work by \citet{healy2026internal} demonstrates that internal states can detect policy tool-selection errors, the Latent Critic targets specification-grounding failures, translating latent uncertainty into localized, natural language feedback.

\textbf{Representation Engineering.} A foundational mechanistic result is that transformer residual streams encode geometrically separable representations of truth and uncertainty \citep{burns2024discoveringlatentknowledgelanguage, marks2024geometrytruthemergentlinear, ferrando2025iknowentityknowledge}. Although interventions like Inference-Time Intervention (ITI) \citep{li2024inferencetimeinterventionelicitingtruthful} and activation patching \citep{meng2023locatingeditingfactualassociations} prove that these latent directions are causally influential, their implementation typically requires sweeping layers or attention heads. Furthermore, generic PEFT does not reliably improve this linear separability and can disrupt the pre-existing geometric structure \citep{hu2026smallupdatesbigdoubts}. Consequently, recent real-time LoRA probes \citep{obeso2026realtimedetectionhallucinatedentities} utilize KL-regularization to explicitly prevent the adapter from altering base dynamics, while adversarial approaches \citep{min2026corvusredteaminghallucinationdetectors} actively collapse separability. The Latent Critic exploits this bidirectional malleability. Drawing structural inspiration from pause tokens \citep{goyal2024thinkspeaktraininglanguage} to accumulate these signals, we design a masked diagnostic objective that restructures the representation space, amplifying the geometric separation of specification grounding to enable an actionable, end-to-end architecture.
\vspace{-10pt}
\section{Conclusion}
\vspace{-5pt}
Our work bridges the gap between mechanistic interpretability and AI agent safety by introducing the Latent Critic. Rather than relying on surface-level text or passive scalar probes, our architecture actively restructures the transformer's residual stream to expose specification-grounding failures in real time. Mechanistically, we demonstrate via causal activation patching and probe analyses that a masked diagnostic objective restructures brittle internal state representations into a more stable, linearly separable subspace. By projecting this amplified geometry into targeted natural language, the Critic provides granular localization with negligible, single-pass latency, outperforming fine-tuned external evaluators, semantic entropy baselines, and passive internal probes. Deploying this architecture in a closed-loop environment proves that agents can utilize their own localized uncertainty as specific, actionable feedback for zero-shot self-correction, mitigating the safety-productivity tradeoff.

Despite these results, several limitations scope our current work. First, training the Critic relies on programmatic labels derived from a simulated environment, and our evaluation sets were verified by a single human annotator. While the programmatic mask provides reliable ground-truth localization for structured tool-calling, extending span-level localization to open-ended generation remains an open challenge. Second, the reliance on the base model's internal state means that not only is downstream self-correction upper-bounded by the frozen model's capabilities, but the internal detection signal itself weakens under severe distribution shift when the base agent's generation quality collapses. Relatedly, we observed that the performance advantage of internal access over external judges under shift is model-dependent. Ultimately, our finding that a masked diagnostic objective restructures latent representations of grounding to improve real-time detection and localization suggests a promising training principle to render internal uncertainty mechanically accessible across transformer architectures through PEFT.
\bibliographystyle{plainnat}
\bibliography{neur}

@misc{vijayvargiya2026ambigsweinteractiveagentsovercome,
      title={Ambig-SWE: Interactive Agents to Overcome Underspecificity in Software Engineering}, 
      author={Sanidhya Vijayvargiya and Xuhui Zhou and Akhila Yerukola and Maarten Sap and Graham Neubig},
      year={2026},
      eprint={2502.13069},
      archivePrefix={arXiv},
      primaryClass={cs.AI},
      url={https://arxiv.org/abs/2502.13069}, 
}

@misc{kim2024aligninglanguagemodelsexplicitly,
      title={Aligning Language Models to Explicitly Handle Ambiguity}, 
      author={Hyuhng Joon Kim and Youna Kim and Cheonbok Park and Junyeob Kim and Choonghyun Park and Kang Min Yoo and Sang-goo Lee and Taeuk Kim},
      year={2024},
      eprint={2404.11972},
      archivePrefix={arXiv},
      primaryClass={cs.CL},
      url={https://arxiv.org/abs/2404.11972}, 
}

@misc{bai2022traininghelpfulharmlessassistant,
      title={Training a Helpful and Harmless Assistant with Reinforcement Learning from Human Feedback}, 
      author={Yuntao Bai and Andy Jones and Kamal Ndousse and Amanda Askell and Anna Chen and Nova DasSarma and Dawn Drain and Stanislav Fort and Deep Ganguli and Tom Henighan and Nicholas Joseph and Saurav Kadavath and Jackson Kernion and Tom Conerly and Sheer El-Showk and Nelson Elhage and Zac Hatfield-Dodds and Danny Hernandez and Tristan Hume and Scott Johnston and Shauna Kravec and Liane Lovitt and Neel Nanda and Catherine Olsson and Dario Amodei and Tom Brown and Jack Clark and Sam McCandlish and Chris Olah and Ben Mann and Jared Kaplan},
      year={2022},
      eprint={2204.05862},
      archivePrefix={arXiv},
      primaryClass={cs.CL},
      url={https://arxiv.org/abs/2204.05862}, 
}

@misc{sun2025llmshallucinateconnectingdots,
      title={Why and How LLMs Hallucinate: Connecting the Dots with Subsequence Associations}, 
      author={Yiyou Sun and Yu Gai and Lijie Chen and Abhilasha Ravichander and Yejin Choi and Dawn Song},
      year={2025},
      eprint={2504.12691},
      archivePrefix={arXiv},
      primaryClass={cs.CL},
      url={https://arxiv.org/abs/2504.12691}, 
}

@inproceedings{ji-etal-2023-towards,
    title = "Towards Mitigating {LLM} Hallucination via Self Reflection",
    author = "Ji, Ziwei  and
      Yu, Tiezheng  and
      Xu, Yan  and
      Lee, Nayeon  and
      Ishii, Etsuko  and
      Fung, Pascale",
    editor = "Bouamor, Houda  and
      Pino, Juan  and
      Bali, Kalika",
    booktitle = "Findings of the Association for Computational Linguistics: EMNLP 2023",
    month = dec,
    year = "2023",
    address = "Singapore",
    publisher = "Association for Computational Linguistics",
    url = "https://aclanthology.org/2023.findings-emnlp.123/",
    doi = "10.18653/v1/2023.findings-emnlp.123",
    pages = "1827--1843"
}

@Article{info16070517,
AUTHOR = {Darwish, Ahmed M. and Rashed, Essam A. and Khoriba, Ghada},
TITLE = {Mitigating LLM Hallucinations Using a Multi-Agent Framework},
JOURNAL = {Information},
VOLUME = {16},
YEAR = {2025},
NUMBER = {7},
ARTICLE-NUMBER = {517},
URL = {https://www.mdpi.com/2078-2489/16/7/517},
ISSN = {2078-2489},
DOI = {10.3390/info16070517}
}

@misc{qwen3.5_2026,
    author = {{Qwen Team}},
    title = {Qwen3.5},
    year = {2026},
    howpublished = {\url{https://qwen.ai/blog?id=qwen3.5}},
    note = {Accessed: 2026-04-13}
}

@misc{suri2026structureduncertaintyguidedclarification,
      title={Structured Uncertainty guided Clarification for LLM Agents}, 
      author={Manan Suri and Puneet Mathur and Nedim Lipka and Franck Dernoncourt and Ryan A. Rossi and Dinesh Manocha},
      year={2026},
      eprint={2511.08798},
      archivePrefix={arXiv},
      primaryClass={cs.CL},
      url={https://arxiv.org/abs/2511.08798}, 
}

@misc{vijayvargiya2025efficientondeviceagentsadaptive,
      title={Efficient On-Device Agents via Adaptive Context Management}, 
      author={Sanidhya Vijayvargiya and Rahul Lokesh},
      year={2025},
      eprint={2511.03728},
      archivePrefix={arXiv},
      primaryClass={cs.AI},
      url={https://arxiv.org/abs/2511.03728}, 
}

@misc{zhang2024xlamfamilylargeaction,
      title={xLAM: A Family of Large Action Models to Empower AI Agent Systems}, 
      author={Jianguo Zhang and Tian Lan and Ming Zhu and Zuxin Liu and Thai Hoang and Shirley Kokane and Weiran Yao and Juntao Tan and Akshara Prabhakar and Haolin Chen and Zhiwei Liu and Yihao Feng and Tulika Awalgaonkar and Rithesh Murthy and Eric Hu and Zeyuan Chen and Ran Xu and Juan Carlos Niebles and Shelby Heinecke and Huan Wang and Silvio Savarese and Caiming Xiong},
      year={2024},
      eprint={2409.03215},
      archivePrefix={arXiv},
      primaryClass={cs.CL},
      url={https://arxiv.org/abs/2409.03215}, 
}

@misc{tang2023toolalpacageneralizedtoollearning,
      title={ToolAlpaca: Generalized Tool Learning for Language Models with 3000 Simulated Cases}, 
      author={Qiaoyu Tang and Ziliang Deng and Hongyu Lin and Xianpei Han and Qiao Liang and Boxi Cao and Le Sun},
      year={2023},
      eprint={2306.05301},
      archivePrefix={arXiv},
      primaryClass={cs.CL},
      url={https://arxiv.org/abs/2306.05301}, 
}

@misc{hu2021loralowrankadaptationlarge,
      title={LoRA: Low-Rank Adaptation of Large Language Models}, 
      author={Edward J. Hu and Yelong Shen and Phillip Wallis and Zeyuan Allen-Zhu and Yuanzhi Li and Shean Wang and Lu Wang and Weizhu Chen},
      year={2021},
      eprint={2106.09685},
      archivePrefix={arXiv},
      primaryClass={cs.CL},
      url={https://arxiv.org/abs/2106.09685}, 
}

@misc{goyal2024thinkspeaktraininglanguage,
      title={Think before you speak: Training Language Models With Pause Tokens}, 
      author={Sachin Goyal and Ziwei Ji and Ankit Singh Rawat and Aditya Krishna Menon and Sanjiv Kumar and Vaishnavh Nagarajan},
      year={2024},
      eprint={2310.02226},
      archivePrefix={arXiv},
      primaryClass={cs.CL},
      url={https://arxiv.org/abs/2310.02226}, 
}

@misc{kadavath2022languagemodelsmostlyknow,
      title={Language Models (Mostly) Know What They Know}, 
      author={Saurav Kadavath and Tom Conerly and Amanda Askell and Tom Henighan and Dawn Drain and Ethan Perez and Nicholas Schiefer and Zac Hatfield-Dodds and Nova DasSarma and Eli Tran-Johnson and Scott Johnston and Sheer El-Showk and Andy Jones and Nelson Elhage and Tristan Hume and Anna Chen and Yuntao Bai and Sam Bowman and Stanislav Fort and Deep Ganguli and Danny Hernandez and Josh Jacobson and Jackson Kernion and Shauna Kravec and Liane Lovitt and Kamal Ndousse and Catherine Olsson and Sam Ringer and Dario Amodei and Tom Brown and Jack Clark and Nicholas Joseph and Ben Mann and Sam McCandlish and Chris Olah and Jared Kaplan},
      year={2022},
      eprint={2207.05221},
      archivePrefix={arXiv},
      primaryClass={cs.CL},
      url={https://arxiv.org/abs/2207.05221}, 
}

@misc{azaria2023internalstatellmknows,
      title={The Internal State of an LLM Knows When It's Lying}, 
      author={Amos Azaria and Tom Mitchell},
      year={2023},
      eprint={2304.13734},
      archivePrefix={arXiv},
      primaryClass={cs.CL},
      url={https://arxiv.org/abs/2304.13734}, 
}

@misc{kossen2024semanticentropyprobesrobust,
      title={Semantic Entropy Probes: Robust and Cheap Hallucination Detection in LLMs}, 
      author={Jannik Kossen and Jiatong Han and Muhammed Razzak and Lisa Schut and Shreshth Malik and Yarin Gal},
      year={2024},
      eprint={2406.15927},
      archivePrefix={arXiv},
      primaryClass={cs.CL},
      url={https://arxiv.org/abs/2406.15927}, 
}

@misc{li2023apibankcomprehensivebenchmarktoolaugmented,
      title={API-Bank: A Comprehensive Benchmark for Tool-Augmented LLMs}, 
      author={Minghao Li and Yingxiu Zhao and Bowen Yu and Feifan Song and Hangyu Li and Haiyang Yu and Zhoujun Li and Fei Huang and Yongbin Li},
      year={2023},
      eprint={2304.08244},
      archivePrefix={arXiv},
      primaryClass={cs.CL},
      url={https://arxiv.org/abs/2304.08244}, 
}

@misc{shinn2023reflexionlanguageagentsverbal,
      title={Reflexion: Language Agents with Verbal Reinforcement Learning}, 
      author={Noah Shinn and Federico Cassano and Edward Berman and Ashwin Gopinath and Karthik Narasimhan and Shunyu Yao},
      year={2023},
      eprint={2303.11366},
      archivePrefix={arXiv},
      primaryClass={cs.AI},
      url={https://arxiv.org/abs/2303.11366}, 
}

@misc{gou2024criticlargelanguagemodels,
      title={CRITIC: Large Language Models Can Self-Correct with Tool-Interactive Critiquing}, 
      author={Zhibin Gou and Zhihong Shao and Yeyun Gong and Yelong Shen and Yujiu Yang and Nan Duan and Weizhu Chen},
      year={2024},
      eprint={2305.11738},
      archivePrefix={arXiv},
      primaryClass={cs.CL},
      url={https://arxiv.org/abs/2305.11738}, 
}

@misc{qin2023toolllmfacilitatinglargelanguage,
      title={ToolLLM: Facilitating Large Language Models to Master 16000+ Real-world APIs}, 
      author={Yujia Qin and Shihao Liang and Yining Ye and Kunlun Zhu and Lan Yan and Yaxi Lu and Yankai Lin and Xin Cong and Xiangru Tang and Bill Qian and Sihan Zhao and Lauren Hong and Runchu Tian and Ruobing Xie and Jie Zhou and Mark Gerstein and Dahai Li and Zhiyuan Liu and Maosong Sun},
      year={2023},
      eprint={2307.16789},
      archivePrefix={arXiv},
      primaryClass={cs.AI},
      url={https://arxiv.org/abs/2307.16789}, 
}

@article{healy2026internal,
  title={Internal Representations as Indicators of Hallucinations in Agent Tool Selection},
  author={Healy, Kait and Srinivasan, Bharathi and Madathil, Visakh and Wu, Jing},
  journal={arXiv preprint arXiv:2601.05214},
  year={2026}
}

@misc{burns2024discoveringlatentknowledgelanguage,
      title={Discovering Latent Knowledge in Language Models Without Supervision}, 
      author={Collin Burns and Haotian Ye and Dan Klein and Jacob Steinhardt},
      year={2024},
      eprint={2212.03827},
      archivePrefix={arXiv},
      primaryClass={cs.CL},
      url={https://arxiv.org/abs/2212.03827}, 
}

@misc{li2024inferencetimeinterventionelicitingtruthful,
      title={Inference-Time Intervention: Eliciting Truthful Answers from a Language Model}, 
      author={Kenneth Li and Oam Patel and Fernanda Viégas and Hanspeter Pfister and Martin Wattenberg},
      year={2024},
      eprint={2306.03341},
      archivePrefix={arXiv},
      primaryClass={cs.LG},
      url={https://arxiv.org/abs/2306.03341}, 
}

@misc{obeso2026realtimedetectionhallucinatedentities,
      title={Real-Time Detection of Hallucinated Entities in Long-Form Generation}, 
      author={Oscar Obeso and Andy Arditi and Javier Ferrando and Joshua Freeman and Cameron Holmes and Neel Nanda},
      year={2026},
      eprint={2509.03531},
      archivePrefix={arXiv},
      primaryClass={cs.CL},
      url={https://arxiv.org/abs/2509.03531}, 
}

@misc{chi2024clarinetaugmentinglanguagemodels,
      title={CLARINET: Augmenting Language Models to Ask Clarification Questions for Retrieval}, 
      author={Yizhou Chi and Jessy Lin and Kevin Lin and Dan Klein},
      year={2024},
      eprint={2405.15784},
      archivePrefix={arXiv},
      primaryClass={cs.IR},
      url={https://arxiv.org/abs/2405.15784}, 
}

@inproceedings{zhang-choi-2025-clarify,
    title = "Clarify When Necessary: Resolving Ambiguity Through Interaction with {LM}s",
    author = "Zhang, Michael JQ  and
      Choi, Eunsol",
    editor = "Chiruzzo, Luis  and
      Ritter, Alan  and
      Wang, Lu",
    booktitle = "Findings of the Association for Computational Linguistics: NAACL 2025",
    month = apr,
    year = "2025",
    address = "Albuquerque, New Mexico",
    publisher = "Association for Computational Linguistics",
    url = "https://aclanthology.org/2025.findings-naacl.306/",
    doi = "10.18653/v1/2025.findings-naacl.306",
    pages = "5541--5558",
    ISBN = "979-8-89176-195-7"
}

@misc{vijayvargiya2026askingmattersrewarddrivenclarification,
      title={Asking What Matters: Reward-Driven Clarification for Software Engineering Tasks}, 
      author={Sanidhya Vijayvargiya and Vijay Viswanathan and Graham Neubig},
      year={2026},
      eprint={2604.14624},
      archivePrefix={arXiv},
      primaryClass={cs.SE},
      url={https://arxiv.org/abs/2604.14624}, 
}

@misc{meng2023locatingeditingfactualassociations,
      title={Locating and Editing Factual Associations in GPT}, 
      author={Kevin Meng and David Bau and Alex Andonian and Yonatan Belinkov},
      year={2023},
      eprint={2202.05262},
      archivePrefix={arXiv},
      primaryClass={cs.CL},
      url={https://arxiv.org/abs/2202.05262}, 
}

@misc{kuhn2023semanticuncertaintylinguisticinvariances,
      title={Semantic Uncertainty: Linguistic Invariances for Uncertainty Estimation in Natural Language Generation}, 
      author={Lorenz Kuhn and Yarin Gal and Sebastian Farquhar},
      year={2023},
      eprint={2302.09664},
      archivePrefix={arXiv},
      primaryClass={cs.CL},
      url={https://arxiv.org/abs/2302.09664}, 
}

@misc{su2024unsupervisedrealtimehallucinationdetection,
      title={Unsupervised Real-Time Hallucination Detection based on the Internal States of Large Language Models}, 
      author={Weihang Su and Changyue Wang and Qingyao Ai and Yiran HU and Zhijing Wu and Yujia Zhou and Yiqun Liu},
      year={2024},
      eprint={2403.06448},
      archivePrefix={arXiv},
      primaryClass={cs.CL},
      url={https://arxiv.org/abs/2403.06448}, 
}

@misc{marks2024geometrytruthemergentlinear,
      title={The Geometry of Truth: Emergent Linear Structure in Large Language Model Representations of True/False Datasets}, 
      author={Samuel Marks and Max Tegmark},
      year={2024},
      eprint={2310.06824},
      archivePrefix={arXiv},
      primaryClass={cs.AI},
      url={https://arxiv.org/abs/2310.06824}, 
}

@misc{ferrando2025iknowentityknowledge,
      title={Do I Know This Entity? Knowledge Awareness and Hallucinations in Language Models}, 
      author={Javier Ferrando and Oscar Obeso and Senthooran Rajamanoharan and Neel Nanda},
      year={2025},
      eprint={2411.14257},
      archivePrefix={arXiv},
      primaryClass={cs.CL},
      url={https://arxiv.org/abs/2411.14257}, 
}

@misc{min2026corvusredteaminghallucinationdetectors,
      title={CORVUS: Red-Teaming Hallucination Detectors via Internal Signal Camouflage in Large Language Models}, 
      author={Nay Myat Min and Long H. Pham and Hongyu Zhang and Jun Sun},
      year={2026},
      eprint={2601.14310},
      archivePrefix={arXiv},
      primaryClass={cs.CR},
      url={https://arxiv.org/abs/2601.14310}, 
}

@misc{hu2026smallupdatesbigdoubts,
      title={Small Updates, Big Doubts: Does Parameter-Efficient Fine-tuning Enhance Hallucination Detection ?}, 
      author={Xu Hu and Yifan Zhang and Songtao Wei and Chen Zhao and Qiannan Li and Bingzhe Li and Feng Chen},
      year={2026},
      eprint={2602.11166},
      archivePrefix={arXiv},
      primaryClass={cs.CL},
      url={https://arxiv.org/abs/2602.11166}, 
}

@article{McNemar_1947, title={Note on the Sampling Error of the Difference Between Correlated Proportions or Percentages}, volume={12}, DOI={10.1007/BF02295996}, number={2}, journal={Psychometrika}, author={McNemar, Quinn}, year={1947}, pages={153–157}}

@misc{beurerkellner2024guidingllmsrightway,
      title={Guiding LLMs The Right Way: Fast, Non-Invasive Constrained Generation}, 
      author={Luca Beurer-Kellner and Marc Fischer and Martin Vechev},
      year={2024},
      eprint={2403.06988},
      archivePrefix={arXiv},
      primaryClass={cs.LG},
      url={https://arxiv.org/abs/2403.06988}, 
}

@misc{manakul2023selfcheckgptzeroresourceblackboxhallucination,
      title={SelfCheckGPT: Zero-Resource Black-Box Hallucination Detection for Generative Large Language Models}, 
      author={Potsawee Manakul and Adian Liusie and Mark J. F. Gales},
      year={2023},
      eprint={2303.08896},
      archivePrefix={arXiv},
      primaryClass={cs.CL},
      url={https://arxiv.org/abs/2303.08896}, 
}

@misc{orgad2025llmsknowshowintrinsic,
      title={LLMs Know More Than They Show: On the Intrinsic Representation of LLM Hallucinations}, 
      author={Hadas Orgad and Michael Toker and Zorik Gekhman and Roi Reichart and Idan Szpektor and Hadas Kotek and Yonatan Belinkov},
      year={2025},
      eprint={2410.02707},
      archivePrefix={arXiv},
      primaryClass={cs.CL},
      url={https://arxiv.org/abs/2410.02707}, 
}

@inproceedings{
patil2025the,
title={The Berkeley Function Calling Leaderboard ({BFCL}): From Tool Use to Agentic Evaluation of Large Language Models},
author={Shishir G Patil and Huanzhi Mao and Fanjia Yan and Charlie Cheng-Jie Ji and Vishnu Suresh and Ion Stoica and Joseph E. Gonzalez},
booktitle={Forty-second International Conference on Machine Learning},
year={2025},
url={https://openreview.net/forum?id=2GmDdhBdDk}
}
\newpage
\appendix

\section{Data Generation and Annotation Pipeline}

To train the Latent Critic, we built an automated, highly scalable pipeline to generate ambiguous tool-calling scenarios and accurately label specification-grounding failures. This avoids the prohibitive cost of human annotation for the large training corpus while maintaining high-fidelity ground truth.

\subsection{Prompt Templates}

We utilize two primary prompts in our pipeline: the Simulated User (which generates the programmatic ground truth mask) and the 122B Oracle Judge (which acts as a final denoising step for complex implicit references).

\begin{tcolorbox}[colback=gray!5,colframe=gray!50,title=Simulated User Prompt, fonttitle=\bfseries]
\small
\begin{verbatim}
You are a user trying to complete tasks. Do not break character.
Persona: {scenario['user_persona']}
Goal: {scenario['scenario_description']}

These are the underlying GROUND TRUTH facts of your request (what you 
eventually want the agent to execute):
{gt_tools}

Based on the agent's responses, behave naturally. If the agent asks for 
missing information, provide it from the Ground Truth facts. Do not instruct 
the agent to call tools or mention the tools. Just respond as a user would, 
providing information when asked.

CRITICAL INSTRUCTIONS:
You MUST output ONLY a valid JSON object. No conversational filler outside 
the JSON. The JSON must contain exactly two keys: "user_message" and 
"completed_specifications".

Your "completed_specifications" must STRICTLY mirror the structure of the 
Ground Truth tools. However, replace the actual values with binary integers 
(0 or 1) representing whether you have PROVIDED that information to the 
agent yet.
1 = You HAVE explicitly or implicitly provided this specific parameter in 
  the conversation.
0 = You HAVE NOT provided it yet.

In your <think> tags, explicitly justify why you are setting each parameter 
to 0 or 1 based on the conversation history. If the agent successfully 
executes the tools and your goal is complete, output exactly: <FINISH>

EXAMPLE OUTPUT FORMAT:
{
 "user_message": "I want to fly to Paris, please.",
 "completed_specifications": [
 {
  "name": "book_flight",
  "arguments": {
  "destination": 1,
  "date": 0,
  "class": 0
  }
 }
 ]
}
\end{verbatim}
\end{tcolorbox}

\begin{tcolorbox}[colback=blue!5,colframe=blue!50,title=122B Oracle Judge Prompt (Denoising), fonttitle=\bfseries]
\small
\begin{verbatim}
You are an expert Data Quality Validator for an AI training pipeline.
The agent just made a tool call. Your job is to read the ACTUAL user 
conversation history and determine if the agent hallucinated.

USER'S CONVERSATION HISTORY:
{history}

AGENT'S TOOL CALL:
{tool_calls}

INSTRUCTIONS:
1. Look at the arguments in the Agent's Tool Call. 
2. Did the User EXPLICITLY state or imply those values in the Conversation 
  History? (e.g., User says "tomorrow", Agent writes "2024-05-16" -> This 
  is grounded).
3. If the agent guessed, assumed, or hallucinated a value the user NEVER 
  provided, the label is 'ungrounded'.
4. If ALL arguments were provided by the user, the label is 'ok'.
5. If arguments can be inferred or are provided implicitly, consider them 
  specified. Specification can be explicit or implicit.

Output ONLY valid JSON:
{
 "thought_process": "Briefly explain.",
 "final_label": "ok" | "ungrounded",
 "specific_errors": "List hallucinated params here, else empty."
}
\end{verbatim}
\end{tcolorbox}

\subsection{Annotation Process Walkthrough}

The classification of a generated tool call into \texttt{ok}, \texttt{wrong\_tool}, or \texttt{ungrounded} is determined through a multi-stage filtering process:

\begin{enumerate}
  \item \textbf{Programmatic Mask Generation (Simulated User):} During the dialogue, the Simulated User maintains a hidden schema matching the underlying API, assigning a binary integer to each parameter: \texttt{1} if the user has provided the information, \texttt{0} if it remains withheld. 
  \item \textbf{Rule-Based Mask Evaluation:} When the base agent generates a tool call, the pipeline first cross-references the API name. If the tool is entirely incorrect for the user's goal, it is labeled \texttt{wrong\_tool}. If the tool is correct, the pipeline checks the generated parameters against the user's binary mask. If the agent outputs a parameter that corresponds to a \texttt{0} in the mask, it is immediately flagged as a potential \texttt{ungrounded} hallucination.
  \item \textbf{Heuristic Fallback Check:} Because LLMs frequently paraphrase (e.g., the user says ``New York'' but the mask expected ``JFK''), a simple heuristic acts as a rapid fail-safe. If the hallucinated parameter's exact string value is present anywhere in the raw user conversation history, the flag is overridden and corrected to \texttt{ok}.
  \item \textbf{122B Oracle Judge Refinement:} For all remaining parameters flagged as \texttt{ungrounded}, the trajectory is passed to an offline 122B LLM Judge (Qwen3.5-122B). The judge performs a deep semantic read to verify if the parameter could be logically inferred or implicitly derived from the context. If it cannot be inferred, the trajectory is definitively labeled \texttt{ungrounded:[parameter\_name]}.
\end{enumerate}

\section{Experimental and Baseline Details}

\subsection{Linear Probe (SVM) Baseline Details}
\label{sec:svm_details}

To establish the internal probing baseline compared in Section 3, we trained layer-wise Support Vector Machines (SVMs) on the hidden states of the base models. For each trajectory in the training set, we extracted the native hidden states of the frozen base model at the final generation token (prior to any adapter intervention). 

An SVM with a linear kernel was trained at each layer to classify the states into grounded (\texttt{ok}) versus hallucinated (\texttt{ungrounded}) tool calls. Through this layer-wise evaluation, we empirically determined that the latent specification-grounding signal peaks in the mid-to-late layers of the network. Specifically, for the \textbf{Qwen3-4B} model, the optimal linear separability was observed at \textbf{Layer 21}. For \textbf{Llama-xLAM-2-8B}, it was observed at \textbf{Layer 31}. All empirical results and baseline metrics reported for the Linear Probes utilize the classification outputs from these optimal layers.

\subsection{External Judge Training Details}
\label{app:external_judge_details}
To ensure a rigorous and strictly controlled comparison between internal representations and external text evaluation, the External Judge baseline is a fine-tuned Qwen3.5-4B model. It was trained on the exact same 16,000 ID examples as the Latent Critic. Furthermore, it utilized identical LoRA hyperparameters (Rank=64, Alpha=128), learning rates, and epochs. Crucially, the External Judge was trained to output the exact same diagnostic classification string format (e.g., \texttt{ungrounded: [parameter\_name]}) based on the CSL-derived labels, allowing us to isolate the impact of \textit{where} the model reads the signal (surface text vs. residual stream) rather than differing model capacities or label distributions.

\subsection{Training Hyperparameters \& Compute}

To ensure reproducibility, the training hyperparameters for the Latent Critic (Parameter-Efficient Fine-Tuning via LoRA) are detailed in Table~\ref{tab:hyperparams}. The base model's weights remain entirely frozen, and gradients are only calculated with respect to the generated \texttt{[POS]} token and its corresponding classification output.

\begin{table}[h]
\centering
\caption{Latent Critic LoRA Training Hyperparameters}
\label{tab:hyperparams}
\begin{tabular}{lc}
\toprule
\textbf{Hyperparameter} & \textbf{Value} \\
\midrule
LoRA Rank ($r$) & 64 \\
LoRA Alpha ($\alpha$) & 128 \\
Target Modules & \texttt{q\_proj, k\_proj, v\_proj, o\_proj} \\
Learning Rate & $2 \times 10^{-5}$ \\
Batch Size & 32 \\
Epochs & 3 \\
Optimizer & AdamW \\
Learning Rate Scheduler & Cosine with Warmup \\
Warmup Ratio & 0.05 \\
Trigger Token & \texttt{[POS]} \\
Loss Masking & Enabled (All context \& JSON syntax masked) \\
\bottomrule
\end{tabular}
\end{table}

\textbf{Training Compute:} Training the Latent Critic and the External Judge involves updating a highly restricted subset of parameters (LoRA adapters). For the Qwen3-4B and Llama-xLAM-2-8B base models, adapter training was performed on a single machine utilizing 1 $\times$ NVIDIA A100 (80GB) GPUs. Due to the aggressive loss masking (calculating gradients strictly for the \texttt{[POS]} token and diagnostic label), training converged rapidly, requiring approximately 3 to 5 GPU-hours per model.

\textbf{Inference Latency Overhead:} A primary contribution of the Latent Critic is its ability to operate without the secondary inference latency typically associated with LLM-as-a-judge frameworks. Because the adapter's forward passes occur concurrently during the base model's standard autoregressive generation, the only added computational overhead is the generation of the final \texttt{[POS]} trigger token and the short diagnostic string (typically 2 to 5 tokens). In standalone HuggingFace testing on an RTX 3090, the Critic added a mean latency of just \textbf{58 ms}, compared to 884 ms for the 4B External Judge and over 12 seconds for 10-sample Semantic Entropy. In our local vLLM deployment environment utilized for closed-loop multi-turn agentic rollouts, this localized single-pass generation added \textbf{$<$10 milliseconds} of latency per executed tool call, safely satisfying the strict real-time requirements of autonomous agents.

\subsection{Architectural Validation (Ablations)}
\label{app:subsec:ablations}
We systematically ablated core components of the architecture to ensure the SFT objective drives maximal gains.

\textbf{Masked Loss Objective:} Removing the tool-call mask and forcing the adapter to optimize for both JSON syntax generation and the diagnostic label resulted in a noticeable 8\% drop in \texttt{ungrounded} detection accuracy. Forcing the adapter to participate in syntactic generation actively dilutes its representation capacity for semantic evaluation and slows convergence.

\textbf{The \texttt{[POS]} Trigger Token:} Removing the trigger token from the training data and forcing the model to seamlessly append the diagnostic label directly to the end of the JSON array caused a severe 22\% drop in accuracy. The adapter cannot effectively evaluate a sequence while simultaneously participating in its construction. The structural boundary of the trigger token is strictly required to act as an attention sink, providing an isolated computational step and acting as a dedicated trigger for presenting its classification.

\section{Detailed Detection Metrics \& Robustness Analyses}
\label{app:full_metrics}

\textit{Note on Metric Thresholds:} In the main text and following tables, the classification F1 scores for the Latent Critic and External Judge are decision-based (i.e., evaluated directly from greedy decoding outputs). In contrast, the F1 scores for the continuous baselines (Linear Probe, Token/Semantic Entropy) are evaluated at the optimal threshold swept over the evaluation split. Thus, the continuous baseline F1 scores are selection-optimistic, emphasizing the necessity of relying on the threshold-independent AUROC and AUPRC metrics reported in the main text for rigorous comparison.

\subsection{Detailed Class-Wise Performance (In-Distribution)}

Table \ref{tab:detailed_class_metrics} provides the granular Precision, Recall, and F1-Scores across all three classes (\texttt{ok}, \texttt{ungrounded}, \texttt{wrong\_tool}) for the primary models evaluated in-distribution. 

\begin{table*}[h]
\small
\centering
\begin{tabular}{ll c ccc c}
\toprule
\textbf{Base Model} & \textbf{Evaluator} & \textbf{Class} & \textbf{Precision} & \textbf{Recall} & \textbf{F1-Score} & \textbf{Param EM (\%)} \\
\midrule
\multirow{9}{*}{\textbf{Qwen-4B}} 
& \multirow{3}{*}{Latent Critic} 
& ok & \textbf{0.874} & 0.937 & \textbf{0.904} & \multirow{3}{*}{\textbf{83.83}} \\
& & ungrounded & \textbf{0.891} & \textbf{0.849} & \textbf{0.870} & \\
& & wrong\_tool & \textbf{0.985} & \textbf{0.930} & \textbf{0.957} & \\
\cmidrule{2-7}
& \multirow{3}{*}{External Judge} 
& ok & 0.706 & \textbf{0.952} & 0.811 & \multirow{3}{*}{70.63} \\
& & ungrounded & 0.857 & 0.585 & 0.695 & \\
& & wrong\_tool & 0.985 & 0.901 & 0.941 & \\
\cmidrule{2-7}
& \multirow{3}{*}{Linear Probe} 
& ok & 0.872 & 0.885 & 0.879 & \multirow{3}{*}{N/A} \\
& & ungrounded & 0.711 & 0.684 & 0.697 & \\
& & wrong\_tool & - & - & - & \\
\midrule
\midrule
\multirow{9}{*}{\textbf{Llama-xLAM 8B}} 
& \multirow{3}{*}{Latent Critic} 
& ok & \textbf{0.966} & 0.875 & \textbf{0.918} & \multirow{3}{*}{\textbf{80.73}} \\
& & ungrounded & \textbf{0.893} & \textbf{0.833} & \textbf{0.862} & \\
& & wrong\_tool & \textbf{0.885} & \textbf{0.979} & \textbf{0.929} & \\
\cmidrule{2-7}
& \multirow{3}{*}{External Judge} 
& ok & 0.962 & 0.781 & 0.862 & \multirow{3}{*}{77.06} \\
& & ungrounded & 0.686 & 0.800 & 0.739 & \\
& & wrong\_tool & 0.875 & 0.894 & 0.884 & \\
\cmidrule{2-7}
& \multirow{3}{*}{Linear Probe} 
& ok & 0.904 & \textbf{0.931} & 0.917 & \multirow{3}{*}{N/A} \\
& & ungrounded & 0.781 & 0.714 & 0.746 & \\
& & wrong\_tool & - & - & - & \\
\bottomrule
\end{tabular}
\caption{Detailed Class-Wise Performance (In-Distribution). $\pm$ uncertainty omitted for layout clarity. The Latent Critic achieves dominant detection across all classes, driving a significant improvement in exact parameter-level match accuracy.}
\label{tab:detailed_class_metrics}
\end{table*}

\subsection{Robustness to Generation Length}

To ensure performance differences are not simply artifacts of length confounding, we evaluated detection performance across equal-count generation-length quartiles (character count), plus the extreme upper decile ($\geq$ P90). The task evaluates binary grounded vs. hallucinated. As detailed in Table \ref{tab:length_robustness}, the Latent Critic consistently outperforms the External Judge across every length stratum.

Furthermore, we computed the correlation between detection errors and output features. None of the examined features strongly predict Critic errors (all $|r| \le 0.091$), whereas the External Judge's errors correlate much more strongly with argument length (Pearson $r = 0.173$, Spearman $\rho = 0.113$).

\begin{table}[h]
\centering
\small
\begin{tabular}{lccc}
\toprule
\textbf{Length Bin (chars)} & \textbf{Base Rate} & \textbf{Latent Critic AUROC} & \textbf{External Judge AUROC} \\
\midrule
$[51, 90)$ & 0.447 & \textbf{0.966 $\pm$ 0.018} & 0.894 $\pm$ 0.042 \\
$[90, 128)$ & 0.608 & \textbf{0.965 $\pm$ 0.023} & 0.926 $\pm$ 0.029 \\
$[128, 174)$ & 0.584 & \textbf{0.980 $\pm$ 0.016} & 0.968 $\pm$ 0.019 \\
$[174, 854)$ & 0.697 & \textbf{0.955 $\pm$ 0.025} & 0.903 $\pm$ 0.040 \\
$\geq 292$ (P90) & 0.645 & \textbf{0.865 $\pm$ 0.080} & 0.803 $\pm$ 0.106 \\
\bottomrule
\end{tabular}
\vspace{5pt}
\caption{Detection performance stratified by output length quartiles. The Critic's advantage holds across all lengths, proving it is not confounded by short vs. long payloads. (Point estimates favor the Critic at P90, though unresolvable at $n=31$).}
\label{tab:length_robustness}
\end{table}

\subsection{Robustness to Task Difficulty}

We utilized two proxies for task difficulty: the number of required tool parameters and the depth of the conversational trajectory (number of turns). As shown in Table \ref{tab:difficulty_strata}, the Latent Critic remains the strongest detector across all levels of task complexity, and its advantage over the External Judge does not diminish under a higher cognitive grounding burden.

\begin{table}[h]
\centering
\small
\begin{tabular}{llcc}
\toprule
\textbf{Difficulty Proxy} & \textbf{Stratum} & \textbf{Latent Critic AUROC} & \textbf{External Judge AUROC} \\
\midrule
\multirow{3}{*}{\# Required Params} 
& 1 parameter & \textbf{0.926} & 0.896 \\
& 2 parameters & \textbf{0.971} & 0.880 \\
& 3+ parameters & \textbf{0.978} & 0.960 \\
\midrule
\multirow{3}{*}{\# Conversational Turns} 
& Depth 1--3 & \textbf{0.959} & 0.897 \\
& Depth 3--5 & \textbf{0.963} & 0.918 \\
& Depth 5--27 & \textbf{0.982} & 0.954 \\
\bottomrule
\end{tabular}
\vspace{5pt}
\caption{Detection performance stratified by task difficulty proxies. Uncertainty margins omitted for layout clarity.}
\label{tab:difficulty_strata}
\end{table}

\subsection{Operating Points and Additional OOD Metrics}
\label{app:operating_points}

To evaluate performance under strict safety constraints, we analyzed the True Positive Rate (TPR) at fixed False Positive Rate (FPR) budgets for the \texttt{ungrounded} class. As shown in Table \ref{tab:operating_points}, the Latent Critic significantly outperforms all baselines in the low-FPR regime. Remarkably, at the strictest 1\% FPR operating point, the Critic's True Positive Rate remains essentially unchanged under distribution shift (38.7\% ID vs. 38.9\% OOD). 

\begin{table}[h]
\centering
\small
\begin{tabular}{l cccc}
\toprule
\multirow{2}{*}{\textbf{Method}} & \multicolumn{4}{c}{\textbf{TPR @ FPR (\%)} (ID)} \\
\cmidrule(lr){2-5}
& \textbf{1\% FPR} & \textbf{5\% FPR} & \textbf{10\% FPR} & \textbf{20\% FPR} \\
\midrule
Token Entropy & 1.9 & 17.0 & 21.7 & 40.6 \\
Semantic Entropy & 4.7 & 13.2 & 32.1 & 41.5 \\
SEP (Kossen et al.) & 0.0 & 0.0 & 10.4 & 22.6 \\
Linear Probe & 3.8 & 20.8 & 41.5 & 66.0 \\
External Judge & 28.3 & 63.2 & 70.8 & 83.0 \\
\textbf{Latent Critic (ID)} & \textbf{38.7} & \textbf{75.5} & \textbf{97.2} & \textbf{98.1} \\
\midrule
\textbf{Latent Critic (OOD)} & \textbf{38.9} & \textbf{68.5} & \textbf{80.6} & \textbf{88.0} \\
\bottomrule
\end{tabular}
\vspace{5pt}
\caption{Operating points (True Positive Rate at specific False Positive Rate budgets) for the \texttt{ungrounded} class on Qwen3-4B. The Latent Critic identifies the most hallucinations under strict safety budgets, and its performance at the strictest 1\% threshold survives distribution shift.}
\label{tab:operating_points}
\end{table}

Additionally, while our primary evaluation focuses on specification grounding, we also tracked performance on the \texttt{wrong\_tool} class under distribution shift. The Latent Critic achieves an OOD F1 of $0.360 \pm 0.120$ compared to the External Judge's $0.087 \pm 0.193$. We interpret this advantage cautiously due to the very small sample size of OOD \texttt{wrong\_tool} instances ($n \approx 16$), but it is consistent with the Latent Critic's overall robustness.

\subsection{Lexical Decidability Analysis}
\label{app:lexical}
A critical consideration is whether our definition of "specification grounding" is simply reducible to verbatim string matching. To test this, we analyzed the presence of parameter string values within the user's conversation history. 

We found that 73.2\% (ID) and 81.3\% (OOD) of grounded parameter values appear verbatim (or after simple normalization) in the dialogue. However, 51.1\% (ID) and 56.7\% (OOD) of \textit{hallucinated} values also appear somewhere in the context. Thus, lexical presence is neither sufficient (as models frequently fabricate parameters by stitching together unrelated context fragments, meaning lexical presence does not equate to being grounded for that specific parameter) nor necessary (as many grounded values, such as enums, booleans, or parsed dates, are implicitly derived rather than copied).

A simple lexical-match baseline achieves an AUROC of 0.670 [0.604--0.731] In-Distribution and 0.731 [0.677--0.783] Out-of-Distribution. While this outperforms the sampling-based uncertainty methods (which collapse entirely), it falls dramatically short of the Latent Critic's 0.966 / 0.925 AUROC. This confirms that accurate detection requires a genuinely semantic, contextual understanding of grounding, rather than simple lexical presence.

\subsection{Out-of-Distribution Failure Taxonomy}
\label{app:ood_taxonomy}

In Section 5.2, we noted that Trajectory Success on the OOD ToolAlpaca schemas remained close to the frozen base model's ceiling ($\sim$2\%) regardless of the safety intervention applied. The ToolAlpaca dataset is substantially more difficult than our ID benchmark. Evaluation trajectories average 12.4 turns and extend up to 32 turns. 

We found that only 27\% of failed evaluation trajectories contain a parameter-level \texttt{ungrounded} or \texttt{wrong\_tool} error. The majority of OOD failures arise from broader capability limitations outside the scope of specification grounding, including malformed tool calls, incorrect tool selection (inventing tools not present in the system prompt), refusal to execute required calls, and schema type errors (e.g., passing strings instead of integers). Often, these failures compound across the long multi-turn trajectories. 

Thus, our contribution under distribution shift is intentionally scoped: when specification-grounding failures \textit{do} occur, the Critic detects them accurately, prevents harmful execution, and enables recovery where possible. However, the ultimate trajectory success rate remains bounded by the base agent's underlying planning and functional tool-calling capabilities.

\section{RQ3: Autonomous Recovery Walkthroughs}
\label{sec:rq3_walkthrough}

To demonstrate the real-world utility of the Latent Critic, we provide a step-by-step qualitative walkthrough of a closed-loop ReAct deployment. This example illustrates how the base model's task-completion bias forces an unsafe hallucination, how the Latent Critic concurrently intercepts the error, and how the injected environmental feedback seamlessly triggers the agent's conversational recovery policy.

\textbf{Scenario Background:} The user requests device management actions (checking battery, enabling power saving, and silencing notifications). The \texttt{manage\_battery} tool requires a \texttt{power\_saving\_mode} parameter (\texttt{medium}, \texttt{maximum}, or \texttt{adaptive}), which the user omits.

\vspace{1em}
\noindent\fbox{%
\parbox{\textwidth}{%
\textbf{1. User Request:} \\
\textit{``Check my battery level, turn on power saving mode, and silence notifications for the next hour.''}
}}

\vspace{1em}
\noindent\fbox{%
\parbox{\textwidth}{%
\textbf{2. Base Agent Generation (Task-Completion Bias Triggered):} \\
The agent recognizes the correct tool but is missing the required mode. Rather than halting to ask, it confidently guesses the missing variable. \\
\texttt{[\{"name": "manage\_battery", "arguments": \{"action": "enable\_power\_saving", "power\_saving\_mode": "medium"\}\}, ...]}
}}

\vspace{1em}
\noindent\fbox{%
\parbox{\textwidth}{%
\textbf{3. Latent Critic Concurrent Extraction (Execution Blocked):} \\
Triggered upon tool call completion, the adapter projects the internal uncertainty directly into the output space without secondary evaluation latency. \\
\textbf{Latent Critic Output:} \texttt{[POS] ungrounded: ["power\_saving\_mode"]}
}}

\vspace{1em}
\noindent\fbox{%
\parbox{\textwidth}{%
\textbf{4. Environmental Feedback Injection:} \\
The execution framework intercepts the tool call based on the Latent Critic's Specific flag and returns a synthesized error to the agent's context window. \\
\textbf{Observation:} \texttt{\{"status": "error", "message": "Parameter mismatch. Expected explicit specification based on user intent, but received 'medium'. Please update the power\_saving\_mode parameter."\}}
}}

\vspace{1em}
\noindent\fbox{%
\parbox{\textwidth}{%
\textbf{5. Agent Conversational Recovery (Self-Correction):} \\
Reading the specific environmental feedback, the base model naturally pivots its policy from autonomous execution to human-in-the-loop clarification. \\
\textbf{Agent:} \textit{``...Power saving mode was set to "medium" instead of what you might have intended. Please update the power saving mode to "maximum" or "adaptive" if you want it set to a specific level.''}
}}

\vspace{1em}
\noindent\fbox{%
\parbox{\textwidth}{%
\textbf{6. User Clarification \& Safe Execution:} \\
\textbf{User:} \textit{``Yes, please set the power saving mode to maximum.''} \\
\textbf{Agent:} \texttt{[\{"name": "manage\_battery", "arguments": \{"action": "enable\_power\_saving", "power\_saving\_mode": "maximum"\}\}]} \\
\textbf{Latent Critic:} \texttt{[POS] ok} \textit{(Execution proceeds safely).}
}}
\vspace{1em}

This trajectory highlights a crucial paradigm shift: the autonomous safety boundary is not enforced by a secondary "judge" model reading surface-level text, but by translating the primary agent's own grounding boundaries into hard environmental constraints. By doing so, we successfully utilize the base model's strong conversational abilities without succumbing to its unsafe execution biases.

\subsection{Failure Mode Walkthroughs}
\label{app:failure_walkthroughs}

To empirically ground the limitations discussed in Section 5.3 and Section 7, we provide two qualitative walkthroughs illustrating trajectories where the Latent Critic successfully intercepts a hallucination, but the base agent fails to achieve ultimate task recovery. These examples highlight why downstream recovery is fundamentally upper-bounded by the base model's native conversational policy. 

\subsubsection{Failure Mode 1: Partial Parameter Recovery (Task Abandonment)}
\textbf{Scenario Background:} The user asks to schedule a meeting, but omits the date, time, and duration. The agent guesses all three. Upon receiving environmental feedback, the agent corrects some parameters but fails to clarify the others, resulting in a continuous block because partial recovery is not rewarded in strict schema execution.

\vspace{2mm}
\noindent\fbox{
\begin{minipage}{\textwidth}
\textbf{1. User Request:} \\
``Schedule a project sync with Sarah.''

\vspace{2mm}
\textbf{2. Base Agent Generation (Task-Completion Bias):} \\
The agent guesses the missing date, time, and duration to force execution. \\
\texttt{[\{"name": "schedule\_meeting", "arguments": \{"contact\_name": "Sarah", "date": "2026-04-28", "time": "14:00", "duration\_minutes": 30\}\}]}

\vspace{2mm}
\textbf{3. Latent Critic Concurrent Extraction (Execution Blocked):} \\
\textbf{Latent Critic Output:} \texttt{[POS] ungrounded: ["date", "time", "duration\_minutes"]}

\vspace{2mm}
\textbf{4. Environmental Feedback Injection:} \\
\textbf{Observation:} \texttt{System Error: Tool blocked. You hallucinated the parameters: "date", "time", "duration\_minutes".}

\vspace{2mm}
\textbf{5. Agent Conversational Pivot (Partial Clarification):} \\
The agent pivots to ask the user, but its native policy forgets to ask for all missing variables.\\
\textbf{Agent:} "What day and time would you like to schedule the project sync with Sarah?"

\vspace{2mm}
\textbf{6. User Response:} \\
\textbf{User:} "Let's do tomorrow at 10 AM."

\vspace{2mm}
\textbf{7. Base Agent Generation (Incomplete Fix):} \\
The agent fills in the date and time, but \textit{still guesses} the duration. \\
\texttt{[\{"name": "schedule\_meeting", "arguments": \{"contact\_name": "Sarah", "date": "2026-04-28", "time": "10:00", "duration\_minutes": 30\}\}]}

\vspace{2mm}
\textbf{8. Latent Critic Concurrent Extraction (Execution Blocked Again):} \\
\textbf{Latent Critic Output:} \texttt{[POS] ungrounded: ["duration\_minutes"]}

\vspace{2mm}
\textbf{9. Final Agent Failure (Task Abandonment):} \\
Rather than asking for the final parameter, the agent's conversational logic degrades. \\
\textbf{Agent:} "I'm sorry, I am having trouble accessing the calendar system right now. Please try scheduling the meeting again later."
\end{minipage}
}

\vspace{4mm}
\subsubsection{Failure Mode 2: Stubborn Guessing Loops}
\textbf{Scenario Background:} The user asks to set an alarm without specifying the time. Instead of routing the error to human-in-the-loop clarification, the agent attempts to blindly guess different parameters to bypass the environmental block.

\vspace{2mm}
\noindent\fbox{
\begin{minipage}{\textwidth}
\textbf{1. User Request:} \\
"Set an alarm for my morning workout."

\vspace{2mm}
\textbf{2. Base Agent Generation (Task-Completion Bias):} \\
\texttt{[\{"name": "set\_alarm", "arguments": \{"time": "06:00", "label": "morning workout", "repeat\_days": ["Monday", "Wednesday", "Friday"]\}\}]}

\vspace{2mm}
\textbf{3. Latent Critic Concurrent Extraction (Execution Blocked):} \\
\textbf{Latent Critic Output:} \texttt{[POS] ungrounded: ["time", "repeat\_days"]}

\vspace{2mm}
\textbf{4. Environmental Feedback Injection (Generic):} \\
\textbf{Observation:} \texttt{System Error: Tool blocked. You hallucinated the parameters: "time", "repeat\_days".}

\vspace{2mm}
\textbf{5. Base Agent Generation (Blind Retry):} \\
Ignoring the implicit instruction to ask the user, the agent immediately tries another plausible guess. \\
\texttt{[\{"name": "set\_alarm", "arguments": \{"time": "07:00", "label": "morning workout", "repeat\_days": ["Everyday"]\}\}]}

\vspace{2mm}
\textbf{6. Environmental Feedback Injection (Generic):} \\
\textbf{Observation:} \texttt{System Error: Tool blocked. You hallucinated the parameters: "time", "repeat\_days".}

\vspace{2mm}
\textbf{7. Final Agent Failure:} \\
\textbf{Agent:} :System error. I cannot set the alarm at this time. What else can I help you with?"
\end{minipage}
}
\vspace{2mm}


\end{document}